\documentclass[lettersize,journal]{IEEEtran}
\usepackage{amsmath,amsfonts}
\usepackage{algorithmic}
\usepackage{algorithm}
\usepackage{array}
\usepackage[caption=false,font=normalsize,labelfont=sf,textfont=sf]{subfig}
\usepackage{textcomp}
\usepackage{stfloats}
\usepackage{url}
\usepackage{verbatim}
\usepackage{graphicx}
\usepackage{amssymb}
\usepackage{booktabs}
\usepackage{multirow}
\usepackage{comment}
\usepackage{xcolor}
\newtheorem{remark}{Remark}

\usepackage{soul}
\usepackage{hyperref}
\usepackage[framemethod=tikz]{mdframed}
\usepackage{mathtools}
\makeatletter
\def\endthebibliography{%
	\def\@noitemerr{\@latex@warning{Empty `thebibliography' environment}}%
	\endlist
}
\makeatother

\begin{document}
\makeatletter
\pretocmd\@bibitem{\color{black}\csname keycolor#1\endcsname}{}{\fail}
\newcommand\citecolor[1]{\@namedef{keycolor#1}{\color{blue}}}
\makeatother 
\title{Synergizing Decision Making and Trajectory Planning Using Two-Stage Optimization for Autonomous Vehicles}

\author{
    Wenru Liu,   
    Haichao Liu,
    Lei Zheng,
    Zhenmin Huang, 
    and Jun Ma
\thanks{This work was supported in part by the National Natural Science Foundation of China under Grant 62303390; in part by the Guangzhou-HKUST(GZ) Joint Funding Scheme under Grant 2024A03J0618. \textit{(Corresponding author: Jun Ma.)}}
    \thanks{Wenru Liu, Haichao Liu, and Lei Zheng are with the Robotics and Autonomous Systems Thrust, The Hong Kong University of Science and Technology (Guangzhou), Guangzhou 511453, China (e-mail: wliu354@connect.hkust-gz.edu.cn; hliu369@connect.ust.hk; lzheng135@connect.ust.hk). }
   \thanks{Zhenmin Huang and Jun Ma are with the Robotics and Autonomous Systems Thrust, The Hong Kong University of Science and Technology (Guangzhou), Guangzhou 511453, China, and also with the Division of Emerging Interdisciplinary Areas, The Hong Kong University of Science and Technology, Hong Kong SAR, China (e-mail: zhuangdf@connect.ust.hk; jun.ma@ust.hk).} 
	}



\maketitle

\begin{abstract}   
   This paper introduces a local planner that synergizes the decision making and trajectory planning modules towards autonomous driving. The decision making and trajectory planning tasks are jointly formulated as a nonlinear programming problem with an integrated objective function. However, integrating the discrete decision variables into the continuous trajectory optimization leads to a mixed-integer programming (MIP) problem with inherent nonlinearity and nonconvexity. To address the challenge in solving the problem, the original problem is decomposed into two sub-stages, and a two-stage optimization (TSO) based approach is presented to ensure the coherence in outcomes for the two stages. The optimization problem in the first stage determines the optimal decision sequence that acts as an informed initialization. With the outputs from the first stage, the second stage necessitates the use of a high-fidelity vehicle model and strict enforcement of the collision avoidance constraints as part of the trajectory planning problem. We evaluate the effectiveness of our proposed planner across diverse multi-lane scenarios. The results demonstrate that the proposed planner simultaneously generates a sequence of optimal decisions and the corresponding trajectory that significantly improves driving performance in terms of driving safety and traveling efficiency as compared to alternative methods. Additionally, we implement the closed-loop simulation in CARLA, and the results showcase the effectiveness of the proposed planner to adapt to changing driving situations with high computational efficiency.
\end{abstract} 


\section{Introduction}
    \IEEEPARstart {A}{utonomous} vehicle (AV) holds the potential to improve driving safety and traveling efficiency in the modern intelligent transportation systems~\cite{chen2022milestones, guo2019safe}.
    The modern AV system typically adheres to a sequential design routine, in which the planning mission can be categorized into three tasks: route planning, decision making, and trajectory planning~\cite{claussmann2019review}.
    Route planning outputs a high-level path based on the road network, while decision making and trajectory planning, collectively referred to as the local planner, focus on the lane-level planning~\cite{teng2023motion}. 
    Safety emerges as the paramount consideration across all tasks undertaken by the AV, whether it pertains to trajectory planning~\cite{wang2022ensuring} or security control~\cite{ding2022security}.
    The imperative of computational efficiency spans from the localization processes~\cite{li2024lidar} involved in route planning to the complexities of trajectory planning~\cite{zheng2023real}. Moreover, the decision making capacity of the AV is significantly influenced by its interactions with other traffic participants, encompassing both passengers~\cite{crosato2022interaction} and other vehicles~\cite{hang2021cooperative}. This consistency in requirements underscores the impetus for adopting a synergistic approach on these individual tasks within autonomous driving~\cite{huang2023differentiable}.
    In practice, unless the AV is approaching its global destination or deviating from the designated route, the responsibility for planning largely falls on the local planner to make decisions (i.e., lane keeping, lane changing, and overtaking), and produce more specific vehicle movements within the lanes~\cite{yurtsever2020survey}. 
    Consequently, integrating the development of decision making and trajectory planning modules into one local planner has received increasing interest, with the motivation to enhance the consistency of outcomes across decision making and trajectory planning tasks{~\cite{sadat2019jointly}}, thereby substantially augmenting the AV's overall driving performance{~\cite{hu2023planning}}.

    This study introduces a local planner that concurrently produces the optimal decision sequences and the corresponding trajectory, aiming to synergize the decision making and trajectory planning modules effectively to enhance the overall driving performance.
    Given that the decision making such as lane-keeping, lane changing, and overtaking is discrete, while trajectory planning is continuous~\cite{huang2024universal}, we employ the mixed-integer programming (MIP) that formulates discrete and continuous variables within a single optimization to precisely model complex scenarios involving the two types of variables.
    With the proposed approach, increased flexibility in optimizing for the driving decisions and the trajectory is ensured, such that it is able to adapt to changing traffic conditions. Also, improved consistency in results for both decision making and planning tasks is attained.
    Specifically, the proposed planner incorporates both decisions and trajectory as optimization variables, and optimizes the decision making and trajectory planning problems using a unified set of planning criteria. The feasibility of the decision making problem is addressed to ensure that decision results are appropriately integrated into the trajectory planning stage through the use of a high-fidelity vehicle model within the local planner.

    Despite the ability of MIP in providing a formulation that aligns well with the inherent nature of the problem in our study, the presence of discrete variable in MIP induces the combinatorial explosion of potential solutions. 
    The solution of MIP problems is classified as NP-hard~\cite{kannan1978computational}, necessitating the deployment of meticulously designed solving algorithms.
    Therefore, a two-stage optimization (TSO) based approach is tailored to achieve high computational efficiency while maintaining coherence for the decision making and trajectory planning outcomes in the proposed local planner. In the first stage, a well-structured MIP problem is formulated and solved for a sequence of optimal decisions. This output serves as an informed initialization for the trajectory optimization in the second stage, ensuring that the results produced in the two stages are coherent, and this leads to a direct improvement in the ultimate solution.
    Our contributions are summarized as follows:
    \begin{enumerate}
        \item A coherent local planner is proposed to enable the AV to make decisions on its optimal reference lane and velocity profile, and the corresponding trajectory is generated to cater to the prevailing traffic scenarios. This eliminates the need for pre-defining the reference lane and velocity profile, and facilitates greater flexibility for the EV in adapting to different driving conditions. 
        \item In the proposed planner, the formulated problem with nonlinearity, nonconvexity, and discrete characteristics is split into two manageable stages so that it can be solved with high computational efficiency. 
        Particularly, the TSO-based approach is introduced to utilize the outcomes derived from the first stage as informed initialization for the second stage, thereby ensuring coherence in the results obtained in both stages.
        \item We empirically evaluate the performance of the proposed coherent local planner through a series of simulations, and the results achieved by the proposed approach demonstrate its superior driving performance compared to alternative methods.
    \end{enumerate}
        
    The organization of this paper is as follows: 
    Section \ref{sec:related} introduces related works on typical decision making and trajectory planning methods.
    Section \ref{sec:pro_form} presents the problem formulation, with the detailed design of the integrated objective function that enables the decision making and trajectory planning problems to be optimized towards the same objective.
    Section \ref{sec:2stage} provides a comprehensive discussion on how the TSO-based approach is utilized to transform the original problem into two sub-stages while maintaining coherence in the results produced.
    Section \ref{sec:exp} presents the experimental results to demonstrate the efficacy of the proposed methodology. 
    The paper concludes with directions for future works in Section \ref{sec:con}.

\section{Related Work} \label{sec:related}
    Decision making and trajectory planning are two core modules in the pipeline of autonomous driving~\cite{paden2016survey}.
    Significant progress has been made in the development of algorithms for decision making~\cite{hubmann2017decision,wang2024chance,cunningham2015mpdm} and trajectory planning~\cite{fan2018baidu,ajanovic2018search,ma2015efficient,chen2019autonomous} in the existing literature.
    However, due to the separation in the design of decision making and trajectory planning modules, each of the two modules may devise its own cost function and give rise to compromise in the quality of the ultimate driving performance~\cite{zhang2021unified}.

    The joint design of the decision making and trajectory planning modules is an emerging trend with the motivation to devise the two modules in pursuit of the same planning objective.
    This effort is bolstered by the application of the learning-based methods.
    For instance, the end-to-end framework in~\cite{bojarski2016end} simply integrates various levels of the planning tasks through a unified neural network, which is further extended by~\cite{codevilla2018end} and~\cite{bansal2018chauffeurnet} to incorporate navigational command and more enriched environmental information into the network.
    However, such a multifaceted system with multiple tasks requires substantial computational resources, which constitute a prevalent constraint of the learning-based approaches. Additionally, the lack of interpretability and theoretical guarantee brings safety concerns.
    In contrast, constrained optimization offers a technique to formulate the problem in an interpretable way, and it scales well if the underlying model aligns well with the problem. 
    However, the decisions, including lane keeping and lane changing maneuvers, are discrete; while the trajectory optimization is generally considered as a continuous-space problem~\cite{schafer2021computation}. 
    The hybrid discrete and continuous characteristics present a challenge in integrating the decision making and trajectory planning problems in the constrained optimization formulation.
    One possible formulation is to work on the continuous aspects of the decision making problem. The decision is assumed to deal with the distribution of the longitudinal velocity profile in~\cite{chen2018continuous}, and is specified as a continuous velocity profile in~\cite{hubmann2016generic}.
    However, the above-mentioned works are limited to decisions on the longitudinal level.
    Relaxation on the discrete decisions is usually applied when extending to the lateral decisions such as lane changing.   
    In~\cite{wang2018predictive}, the discrete maneuvering decisions are integrated into an MPC-based motion planning scheme, and solved by relaxation.
    A unified MPC that automatically decides the mode of maneuvers and the path planning is proposed in~\cite{liu2017path}, and the MIP is turned into a nonlinear programming problem using convex relaxation.  
    Though with computational tractability, such relaxation causes the vehicle to straddle multiple lanes as the vehicle is not confined to committing to a single lane~\cite{wang2018predictive}, and the completion of driving  relies heavily on the progression towards the pre-defined destination~\cite{liu2017path}, therefore lacking applicability to discretionary driving scenarios.

    Opposite to the relaxation of the integer variable, integrating the discrete decision making variables in the optimization problem requires solving of MIP problems, with typical algorithms including branch-and-bound, branch-and-cut, and cutting-plane methods~\cite{boyd2007branch}. These algorithms rely on strategic reduction and refinement of the solution space to efficiently find the optimal solution.
    In the context of autonomous driving, the research involving MIP poses higher requirements in real-time performance, so approximations and heuristic methods are often leveraged.
    Simplified vehicle models are utilized as an approximation of real vehicle model in ~\cite{qian2016optimal, esterle2020optimal} to alleviate the computation burden. But the approximation induces model mismatch with the actual system, thereby impeding the applicability of the results as noted by~\cite{wang2020infusing}.
    To avoid simplifications in the above works, a two-stage approach utilizes a preliminary stage, typically a discrete method, to tackle the decision making problem first as heuristic. The integrated decision making and trajectory planning problem in ~\cite{ammour2022mpc} relies on a finite state machine (FSM) to select an appropriate driving task in the first stage, which is subsequently formulated as the constraints in a trajectory planner in the second stage.
    In~\cite{hang2020integrated}, game theory is firstly applied to model the decision making process, and then an optimizer is leveraged to predict the states of the AVs. 
    In~\cite{fisac2019hierarchical}, the formulated problem is decomposed into a strategic game and a tactical game, where the simplified strategic game is solved in the first place.
    When the planning problem is integrated with the discrete rule hierarchies, the coarse trajectory is selected first and passed to a continuous optimization for further refinement~\cite{veer2023receding}.
    The utilization of the two-stage approach manages to solve the problem, but the consistency of the results attained in the preliminary stage and the ultimate solution is not always guaranteed since distinct methods are utilized in the two stages. 
    The TSO-based approach~\cite{eiras2021two}, on the other hand, allows the first stage to take a similar problem representation as the original problem, enabling the solution from the first stage to act as an informed initialization for the second stage.
    Although the original problem is split into two sub-stages, the utilization of the first stage can ensure the quality of the ultimate solution. 
    Nevertheless, it essentially addresses the motion planning problem, while our work focuses on the joint design of the decision making and trajectory planning modules as a coherent local planner.     

\section{Problem Formulation} \label{sec:pro_form}
    \begin{figure}[t]
		\centering
		\includegraphics[trim=0 0 0 0, clip, width=1\linewidth]{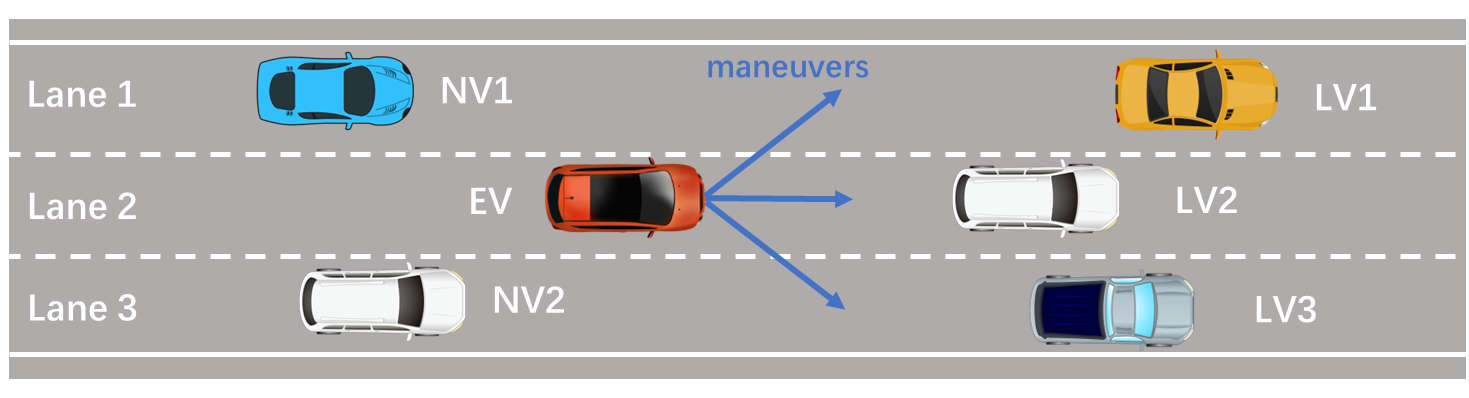}
		\caption{The ego vehicle (EV) in red is driving in a multi-lane scenario accompanied by various surrounding vehicles (SVs). The SVs are further labeled as the leading vehicles (LVs) in front of the EV and the neighbor vehicles (NVs) approaching from behind.}
		\label{fig:lc_demo}
	\end{figure}
    As shown in Fig.~\ref{fig:lc_demo}, we consider a multi-lane scenario consisting of the ego vehicle (EV) and multiple surrounding vehicles (SVs). 
    Note that the SVs are further classified into the leading vehicles (LVs) and the neighbor vehicles (NVs) according to their relative positions to the EV.

    In this study, we make the following assumptions:
    \begin{enumerate}
        \item Discretionary driving: A global path at the road level is accessible; however, the EV is not assigned to adhere to a predefined reference lane. Instead, it is tasked with lane-level planning, which encompasses selecting the reference lane for travel and generating the corresponding trajectory for execution. The primary goal of this planning is to enhance safety by improving driving conditions, such as transitioning to lanes with fewer SVs, or enhancing travel efficiency by switching lanes to achieve higher velocities.
        \item Lane order: Each lane is affiliated with a lane ID. The leftmost lane is considered as lane $1$, and the lane ID is incremented by one from the left to right.
        \item Velocity regulation: The velocity of the EV is governed by the velocity limit of the travel lane. Furthermore, the EV is expected to sustain its velocity within the limit of the vehicle directly preceding it for safety concerns.
        \item Full access to the lane and prediction information: The lane information and the predicted future states of the LVs and NVs, including positions and velocities, are known to the EV.
    \end{enumerate}

\begin{figure}[t]
        \centering
        \includegraphics[trim=0 0 0 0, clip, width=1\linewidth]{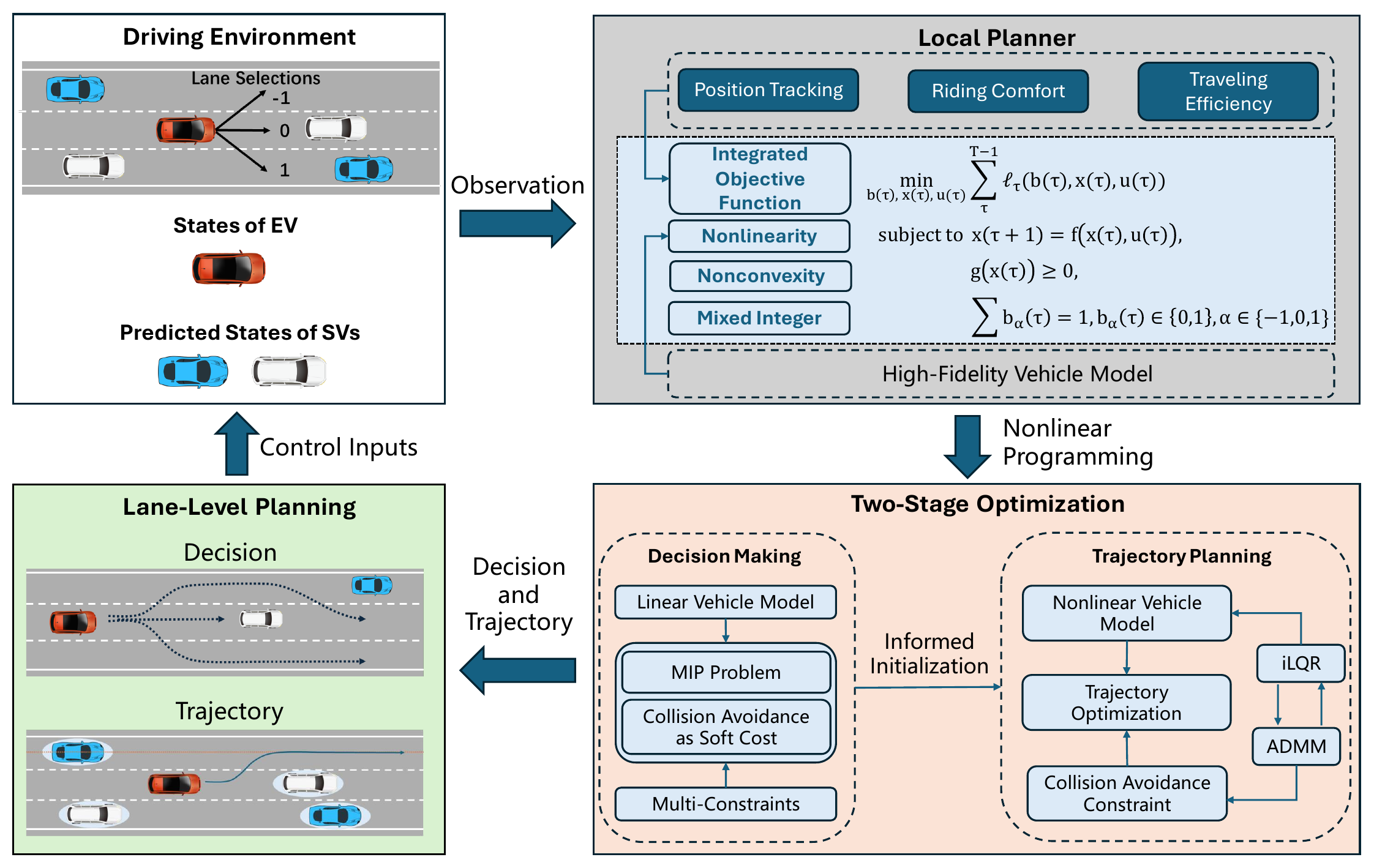}
        \caption{Overview of the proposed planner for autonomous driving. We cast the decision making and trajectory planning modules into a nonlinear programming formulation. The two modules are jointly designed using an integrated objective function, and the TSO-based approach is proposed to solve the formulated problem in two stages. We first propose an MIP formulation to determine the optimal decision sequences, and then solve a well-defined trajectory optimization problem to generate motion plans for the EV.}
        \label{fig:framework}
    \end{figure}

    The driving framework is shown in Fig.~\ref{fig:framework}.
    The EV is able to access the environment information and the global path provided by the router. 
    The focus of this study is the decision making problem (i.e., car following, lane changing, and overtaking), and the trajectory planning problem. The two modules are designed coherently as a local planner to fulfill the planning task at the lane level, wherein the outputs of the local planner are an optimal sequence of decisions and respective trajectory at the same time. 
    The core of the coherent local planner is a two-stage optimization, where the decision making and trajectory planning tasks share similar considerations in the design of the planning objective. 
    The decision making process considers the interaction of the EV with the SVs to provide a guidance towards solution with less influences from the SVs, which leads to enhanced safety and higher traveling efficiency.
    The trajectory planning further necessitates the hard enforcement of the collision avoidance constraint and the utilization of the high-fidelity vehicle model to generate a high-quality trajectory.
    
    \subsection{Vehicle Model} \label{sec:nonlinear_vm}
        In our study, we utilize the bicycle vehicle model~\cite{huang2023decentralized, liu2024improved} to represent the dynamics of the EV, opting for a simplified two-wheel framework to enhance manageability. This model is predicated on several underlying assumptions, including the vehicle's operation on a flat surface, the rigidity of the vehicle's structure, and the existence of a linear correlation between tire forces and slip angles.
        We define the EV's state with $x=(p_x, p_y, \theta, v)$, where $p_x$ and $p_y$ are the positions of the center point of the vehicle on the $X$ (longitudinal) and $Y$ (lateral) directions, respectively; $\theta$ is the heading angle of the vehicle; and $v$ is velocity of the vehicle.
        We further define the EV's control inputs with $u(\tau) = (\delta, a)$, where $a$ and $\delta$ are the acceleration and steering angle, respectively. 
        
        With $\tau_s$ being the time step, we have the following expression for the nonlinear discrete vehicle model:
        \begin{equation} \label{eq:dynamics}
            \begin{aligned}
            p_x(\tau+1) &= p_x(\tau)+f_r(v(\tau),\delta(\tau))\cos(\theta(\tau)), \\
            p_y(\tau+1) &= p_y(\tau)+f_r(v(\tau),\delta(\tau))\sin(\theta(\tau)), \\
            \theta(\tau+1) &= \theta(\tau)+\arcsin(\frac{\tau_s v(\tau)\sin(\delta(\tau))}{h}), \\
            v(\tau+1) &= v(\tau)+\tau_sa(\tau).
            \end{aligned}
        \end{equation}
        
        The function $f_r(v, \delta)$ is defined as
        \begin{equation} \label{fr}
            \begin{aligned}
            f_r(v, \delta) &= h + \tau_sv\cos(\delta) - \sqrt{h^2-(\tau_sv\sin(\delta))^2},
            \end{aligned}
        \end{equation}
        where $h$ denotes wheelbase of the vehicle.

    \subsection{Integrated Objective Function}
        A significant improvement of a coherent local planner is that the decision making and trajectory planning tasks are optimized toward the same goal. The integrated objective function provides the basis for this motivation.
        We start with a concrete definition of the decisions in this work.
        Essentially, the decisions for the AVs take place in both the lateral and longitudinal directions. 
        The lateral decision refers to the selection of the reference lane, where the center-line of the reference lane is usually provided in the lane-centric method.
        The changing of the reference lane manifests decision actions like lane keeping, lane changing, and overtaking (i.e., the reference lane changes and then reverts back).
        The decision in the longitudinal direction is to track certain velocity profiles. By adjusting its velocity, the AV is able to fulfill actions like car following, stopping, or accelerating to overtake.
        With the above discussion, the decision in this work is defined as lane selection, which directly determines the reference lane and velocity profile. 

        The lane selection decision $b$ is defined with a set of binary variables $b_{\alpha}$, where $\alpha \in \{{-1},0,1\}$. Notably, $-1$, $0$, and $1$ in this set denote the left lane change, lane keeping, and right lane change maneuvers, respectively. $b_{\alpha}$ equals to one if the value of $\alpha$ is selected and equals to zero otherwise. 
        As the EV can only select one decision at one time, we further include a constraint to imply that the sum of the values of $b_{\alpha}$ over all possible values of $\alpha$ must equal to one. Subsequently, the complete definition on the binary variable $b_{\alpha}$ is given as follows:
            \begin{align}
            \sum_{\alpha} b_{\alpha}(\tau) = 1, b_{\alpha}(\tau) \in \big\{0,1\}, \alpha \in \{{-1},0,1\}. 
            \end{align}
        
        It is worth noting that, according to the lane ID in this study, we introduce the concept of the target lane with the following definition:
        Suppose the EV is initially in lane $\sigma$ and the decision is $\alpha$, the target lane is thereby lane $\sigma+\alpha$.     
    
        Unlike the discrete nature of the decision making problem, the trajectory planning problem is generally optimized in the continuous space. 
        Considering the discrete-time setting, the trajectory $\eta$ is defined as a sequence of the EV's states at discrete time steps. 
        The objective function over $T$ discrete time steps is then defined as:
        \begin{align}\label{}
            J = \sum_{\tau=0}^{T-1} \ell_\tau\big(b(\tau), x(\tau),u(\tau)\big).
        \end{align}    

        The integrated objective function is constructed on the basis of the general formulation of the objective function in the trajectory optimization, including the position tracking, longitudinal velocity tracking, and riding comfort costs~\cite{ma2022alternating}.
    
        The position tracking cost is defined as the distance between the EV's actual positions and the coordinates of the center-line of the reference lane. It is a necessary component in the integrated function of the local planner because adhering to the geometric structure of the road is a requirement in common to both the decision making and trajectory planning tasks.
        Different from the pipeline where the decision is made individually and the trajectory planning is provided with the reference lane as a pre-defined tracking target, the decision making is part of the optimization process. The exact value of the reference lane is not available in advance. Instead, each decision result gives rise to a different reference lane, and consequently, a different position tracking cost. 
        The position tracking cost is devised to be directly associated with each decision result $\alpha$:
            \begin{align}\label{}
                &\ell_{tr,\alpha}(\tau) = w_{tr,x} (p_{x}(\tau) - p_{x,\sigma+\alpha}^r(\tau))^2  \\
                &+ w_{tr,y} (p_{y}(\tau) - p_{y,\sigma+\alpha}^r(\tau))^2 \nonumber,
            \end{align}
            where $p_{x,\sigma+\alpha}^r$ and $p_{y,\sigma+\alpha}^r$ are the $X$ and $Y$ coordinates of the center of the target lane $\sigma+\alpha$ with respect to the decision $\alpha$, and the positive constants $w_{tr,x}$ and $w_{tr,y}$ are the weighting coefficients.
      
        The position tracking cost in the integrated function is defined as the following, taking into consideration all possible values of the decision $\alpha$:
        \begin{align} \label{eq:l_bxu}
           & \ell_{\tau, tr}\big(x(\tau),u(\tau),b(\tau)\big)   \\
           &= b_{-1}(\tau) \ell_{tr,-1}(\tau) 
           + b_{0}(\tau) \ell_{tr,0}(\tau) 
           + b_{1}(\tau) \ell_{tr,1}(\tau). \nonumber 
        \end{align}

        Similar to the position tracking cost, the longitudinal velocity tracking cost aims to ensure that the AV progresses towards a target velocity. Again, the reference longitudinal velocity is no longer provided as the input to the proposed local planner. The AV relies on the optimization process to determine the reference velocity to track.
        In determining the target velocity to track, the EV must adhere to the velocity limit and should not travel faster than the vehicle directly ahead in the same lane.
        Nevertheless, the EV can achieve a higher velocity by changing to adjacent lanes, which provides the rationale for decisions like lane changing and overtaking.
        Based on the decisions $\alpha$ and the target lane $\sigma+\alpha$, {the velocity limit of the target lane and the velocity of the LV can be derived, which directly determines the target velocity. 
        The common objective in the discretionary driving considered in this study is to attain higher velocity in addition to purely tracking any pre-defined velocity. In turn, the longitudinal velocity tracking cost is renovated as the traveling efficiency cost, which is defined as follows:               
            \begin{align}\label{}
                \ell_{eff,\alpha}(\tau)= w_{velo} (v_{x}(\tau) - v_{x,\sigma+\alpha}^{r}(\tau))^2,
            \end{align}
            where $w_{velo}$ is the weighting parameter, and $v_{x,\sigma+\alpha}^r(\tau)$ is the reference velocity on the target lane $\sigma+\alpha$:
            \begin{align}\label{}
                v_{x,\sigma+\alpha}^r(\tau) = \min(v_{x,\sigma+\alpha}^{limit}(\tau),v_{x,\sigma+\alpha}^{LV}(\tau)),
            \end{align}
        with {$v_{x,\sigma+\alpha}^{limit}$} as the velocity limit of the target lane, and {$v_{x,\sigma+\alpha}^{LV}$} as the velocity of the LV.

        The traveling efficiency cost in the integrated objective function, considering each decision value, is defined as follows:
            \begin{align} \label{}
               & \ell_{\tau, eff}\big(x(\tau),u(\tau),b(\tau)\big) 
               = b_{-1}(\tau) \ell_{eff,-1}(\tau)  \\ 
                &+ b_{0}(\tau) \ell_{eff,0}(\tau) 
                + b_{1}(\tau) \ell_{eff,1}(\tau) \nonumber. 
            \end{align}

        We further supplement the construction of the integrated function with the riding comfort cost defined as follows:
            \begin{align}\label{}
                \ell_{\tau, rc}
                = w_{\delta} (\delta(\tau))^2 + w_{a} (a(\tau))^2,
            \end{align}
        where the positive constants $w_{\delta}$ and $w_{a}$ are weighting coefficients.

        Overall, the objective function is defined as follows:
        \begin{align} \label{eq:obj_basis}
           & \ell_\tau\big(b(\tau),x(\tau),u(\tau)\big)
            =  \ell_{\tau, tr} + \ell_{\tau, eff} + \ell_{\tau, rc}. 
        \end{align}

        \begin{remark}
            The integrated objective function provides the basis for both the decision making and trajectory planning problems. It is constructed upon common considerations in both modules, ensuring that they are optimized towards the same goal.
            It is worth noting that there exists a mixed integer expression in the integrated objective function, which originates from the hybrid nature of the discrete decision making problem and continuous trajectory planning problem.
        \end{remark}
            
    \subsection{Nonlinear Programming Problem} \label{sec:NLPP}
        With the integrated objective function, the proposed local planner can be formulated as a nonlinear programming problem as follows:
        \begin{subequations} 
        \begin{align} 
             \displaystyle\operatorname*{minimize}_{b(\tau),x(\tau),u(\tau)}\quad
            &  \sum_{\tau=0}^{T-1} \ell_\tau\big(b(\tau), x(\tau),u(\tau)\big) \label{eq:2a} \\
            \operatorname*{subject\ to}\quad
            &x(\tau+1)=f\big(x(\tau),u(\tau)\big)  \label{eq:2b}, \\
            &x(\tau)\in X, u(\tau)\in U  \label{eq:2c}, \\
            &g(x(\tau))\ge0  \label{eq:2d}, \\ 
            &x(0)=x_0  \label{eq:2e}, \\
            &\sum_{\alpha}b_{\alpha}(\tau) = 1, \label{eq:alpha1}\\
            & b_{\alpha}(\tau) \in \big\{0,1\}, \label{eq:alpha2}\\
            & \alpha \in \{{-1},0,1\}, \label{eq:alpha3}  \\
            &\tau=0,1,\dotsm, T-1. \nonumber
        \end{align}   
        \end{subequations}

        Note that \eqref{eq:2a}} is the integrated objective function;
        \eqref{eq:2b} is the constraint imposed by the vehicle model;
        \eqref{eq:2c} represents the state and input constraints;
        \eqref{eq:2d} is the constraint for collision avoidance;
        \eqref{eq:2e} represents the initial states of the system.
        We further suffice the problem with additional constraints \eqref{eq:alpha1}, \eqref{eq:alpha2}, and \eqref{eq:alpha3} because of the introduction of the binary variables.
    
        The above optimization problem presents the following challenges: 
        \begin{enumerate}
            \item Nonlinearity: Generating a dynamically feasible trajectory necessitates a high-fidelity nonlinear dynamic model. This introduces a significant challenge in solving for a feasible trajectory considering the inherent characteristics of nonlinearity. 
            \item Nonconvexity: Ensuring the safety of the EV in the presence of multiple SVs involves nonconvex collision avoidance constraints, which increases the computational burden of solving this nonlinear programming problem.
            \item Hybrid discrete and continuous characteristics: Integrating discrete decisions, characterized by abrupt changes in behavior such as lane keeping and lane changing maneuvers, into a continuous trajectory optimization framework poses a significant challenge. Bridging this gap between discrete and continuous domains requires innovative solutions.
        \end{enumerate}

        Typically, it is computationally intractable to directly solve this optimization problem {\eqref{eq:2a}}-{\eqref{eq:alpha3}} with algorithms such as branch-and-bound, or optimization solving mechanisms like iterative linear quadratic regulator (iLQR).
        The underlying challenge in practice is that the existing MIP algorithms are not catered to systems with complex nonlinear model as in \eqref{eq:2b}, and general optimization solvers cannot effectively handle the integer programming aspects. 
        In view of the aforementioned difficulties, this work divides the original problem into manageable sub-stages. This approach combines the merits of existing techniques in both MIP and general optimization, leading to the TSO-based approach developed in the next section

\section{Two-Stage Optimization-Based Approach} \label{sec:2stage}
    In this section, we leverage the TSO-based approach~\cite{eiras2021two} to solve the nonlinear programming problem formulated as above. The TSO-based approach aligns with our motivation to decompose the original problem into two sub-stages while ensuring the coherence of the two optimization stages within the local planner.      
    The TSO-based approach is innovated in this study as the problem in focus is different.
     
    In the first stage, the nonlinear programming is transformed into a well-formulated MIP problem targeting at solving for the discrete lane selection variable $b$. 
    Using the inputs from the first stage, the second stage is formulated as a general trajectory planning problem.
    The problem transformations regarding the different stages are two folds:
    Firstly, a linear vehicle model is utilized to enhance the computational efficiency in optimizing the decisions, and a high-fidelity vehicle model is used in the next stage to generate a feasible trajectory. We follow the work in~\cite{eiras2021two} for the usage of a linear vehicle model as a proxy for the high-fidelity vehicle model.
    Secondly, the collision avoidance constraint is adapted as the soft constraint in the MIP problem in the first stage, while a strict requirement regarding collision avoidance is further imposed in the second stage. 

    In this way, the first stage (see Section \ref{sec:two_mip}) solves an MIP problem to obtain the lane selection decision, from which the reference lane and reference velocity can be derived. 
    Next, the second stage (see Section \ref{sec:ilqr}) solves a trajectory planning problem based on the outputs from the first stage.
    As the problem representations are similar in the two stages, the first stage acts as a proxy to the overall problem, and the two stages are optimized towards the ultimate solution.

    \subsection{Precursory Optimization on Decision Making} \label{sec:two_mip}
        \subsubsection{Linear Vehicle Model} \label{sec:simple_vm}
            We consider a double integrator vehicle model, with the state vector $\bar{x} = (p_x, p_y, v_x, v_y)$ and the control input $\bar{u} = (a_x, a_y)$, where
                \begin{IEEEeqnarray}{rCl}
                    v_x = v\text{cos}(\theta), \quad v_y = v\text{sin}(\theta).  \IEEEeqnarraynumspace
				\end{IEEEeqnarray}
            
            The state-space representation of the double-integrator vehicle model in continuous time is as follows:
            \begin{align} \label{eq:vm_continuous}
                \dot{\bar{x}}(t) = A {\bar{x}}(t) + B {\bar{u}}(t),
            \end{align}
            where the system matrix $A$ and the input matrix $B$ are defined as:
        		\begin{IEEEeqnarray}{rCl}
					A
                     =\left[\begin{array}{c}
					0 \quad 0 \quad 1 \quad 0 \\
					0 \quad 0 \quad 0 \quad 1 \\
                    0 \quad 0 \quad 0 \quad 0 \\
                    0 \quad 0 \quad 0 \quad 0 \\
					\end{array}\right], \
                    B=
                    \left[\begin{array}{c}
					0 \quad 0  \\
					0 \quad 0  \\
                    1 \quad 0  \\
                    0 \quad 1  \\
					\end{array}\right] . \IEEEeqnarraynumspace
				\end{IEEEeqnarray}

            Then, the discrete-time linear vehicle model is defined as
                \begin{IEEEeqnarray}{rCl} \label{eq:simple_vm}
                    \bar{x}(\tau+1) = F(\bar{x}(\tau), \bar{u}(\tau)) = A_d\bar{x}(\tau) + B_d\bar{u}(\tau), \IEEEeqnarraynumspace
				\end{IEEEeqnarray}  
            where $F$ is a zero-order hold discretization of the continuous state-space system \eqref{eq:vm_continuous}. 
            With the time step $\Delta t$, $A_d$ and $B_d$ are defined as:
            
            \begin{IEEEeqnarray}{rCl}
            A_d = 
            \begin{bmatrix} 
            1 & 0 & \Delta t & 0 \\ 
            0 & 0 & 0 & \Delta t \\
            0 & 0 & 1 & 0 \\ 
            0 & 0 & 0 & 1  
            \end{bmatrix}, \
            B_d=        
            \begin{bmatrix} 
            0 & 0  \\ 
            0 & 0  \\
            1 & 0  \\ 
            0 & 1  
            \end{bmatrix}.
            \end{IEEEeqnarray}

        The utilization of this linear vehicle model is a way to introduce linearity in the problem formulation as compared to the original formulation with the vehicle model which is high-fidelity but nonlinear. Noting that the linear vehicle model is simply and is only an approximation of the original vehicle model, the solution from our first stage acts as a proxy, but is not of kinematic feasibility.
        
        \subsubsection{Constraints}
            Following the discussion on the linear vehicle model, the following constraint is imposed to approximate kinematic feasibility, with a constant $\rho \in \mathbb R^{+}$. 
                \begin{IEEEeqnarray}{rCl} \label{eq:simple_vm_con}
                    v_x \ge \rho \mid{v_y}\mid. \IEEEeqnarraynumspace
				\end{IEEEeqnarray}
            Note that this constraint regulates the consistency in the movement along the $X$ and $Y$ axes, to ensure a forward motion of the vehicle. Additionally, the velocity is also constrained with $v_{x,\text{min}} \geq 0$ to guarantee forward motion. 

            The same constraints as in the nonlinear bicycle model \eqref{eq:dynamics} are considered. The input constraints taking into account the engine force limit and braking force limit of the vehicle are shown as:
            \begin{align}
                a_{x,\text{min}} &\leq a_x(\tau) \leq a_{x,\text{max}}, \label{eq:ax} \\
                a_{y,\text{min}} &\leq a_y(\tau) \leq a_{y,\text{max}}, \label{eq:ay}
            \end{align}
            where $a_{x,\text{max}}$ and $a_{y,\text{max}}$ are the maximum values of the acceleration, which are determined by the engine force limit; and $a_{x,\text{min}}$ and $a_{y,\text{min}}$ are the minimum values of the acceleration, which are determined by the braking force limit. 

        \subsubsection{Cost Function} \label{sec:soft}
        It is essential for the cost function in the first stage to be similar to the one in the nonlinear programming problem, thereby ensuring a minimal disparity between the optimal solutions of the two stages. 
        
        Based on \eqref{eq:obj_basis}, two soft constraints are introduced to consider the impact of the LVs and NVs on the EV, since the interactions between vehicles can be viewed as a coupling cost~\cite{kavuncu2021potential}. The collision avoidance constraint \eqref{eq:2d} is transformed to the soft constraint to facilitate efficient decision making in the first stage, utilizing the decision making model derived from driving behavior analysis of human drivers{~\cite{hang2020human}}.

        We first examine the cost associated with the LVs. In lane-keeping and lane-changing scenarios, if an LV is present on the target lane, the EV needs to keep a safe longitudinal distance from that LV.
        Driving too close to the LV or traveling faster than the LV introduces potential hazards to driving safety. 
        Considering the decision $\alpha$ and the respective target lane $\sigma+\alpha$,         we denote the positions of the LV on the target lane as $p_{x,\sigma+\alpha}^{LV}$ and $p_{y,\sigma+\alpha}^{LV}$, and the longitudinal velocity of the LV as $v_{x,\sigma+\alpha}^{LV}$.
        The cost term $\ell_{d-lon,\alpha}$ is defined to describe the cost arising from the interaction between the EV and the LV for the respective decision $\alpha$, and we have
        \begin{align}\label{eq:lon-cost}
            \ell_{d-lon,\alpha}(\tau) &= w_{v-lon} \xi_{\sigma+\alpha}^{LV}(\tau) ( v_{x,\sigma+\alpha}^{LV}(\tau)-v_{x}(\tau))^2  \\
            & + \frac{w_{d-lon}}{(\Delta d_{\sigma+\alpha}^{LV}(\tau))^2+\varepsilon^2}, \nonumber
        \end{align}
        \begin{align}\label{eq:lon-d}
            & \Delta d_{\sigma+\alpha}^{LV}(\tau)  \\
            & = \sqrt{(p_{x,\sigma+\alpha}^{LV}(\tau)-p_{x}(\tau))^2+(p_{y,\sigma+\alpha}^{LV}(\tau)-p_{y}(\tau))^2} - l, \nonumber
        \end{align}
        \begin{align}\label{eq:lon-sgn}
            \xi_{\sigma+\alpha}^{LV}(\tau)= \left\{ \begin{aligned}
                &1\quad\quad \text{if} \ v_{x,\sigma+\alpha}^{LV}(\tau)-v_{x}(\tau)<0 \\
                &0 \quad\quad \text{if} \ v_{x,\sigma+\alpha}^{LV}(\tau)-v_{x}(\tau) \geq0, \\
            \end{aligned}
            \right.
        \end{align}
        where the positive constants $w_{v-lon}$ and $w_{d-lon}$ are weighting coefficients, and a small nonzero constant $\varepsilon$ is added to the denominator to prevent the denominator from being zero;
        $l$ is the length of the vehicle; 
        $\xi^{LV}$ is a switching function such that the difference in velocities is penalized only if the velocity of the LV is lower than the EV's, which is reasonable because the safe distance is only relevant when there is a slow-moving vehicle ahead. 

        By considering all possible values of $\alpha$, we have the soft cost on the LVs as follows:
        \begin{align} \label{eq:l_bxu}
           &\ell_{\tau, s\_LV}(\tau) = b_{-1}(\tau) \ell_{d-lon, -1}(\tau)  \\
           & + b_{0}(\tau) \ell_{d-lon, 0}(\tau) + b_{1}(\tau) \ell_{d-lon, 1}(\tau).  \nonumber
        \end{align}
        Similarly, we consider the cost associated with the NVs, defined as:
        \begin{align}\label{eq:lat-cost}
            \ell_{d-lat,\alpha}(\tau) &=  w_{v-lat} \xi_{\sigma+\alpha}^{NV}(\tau) ( v_{x,\sigma+\alpha}^{NV}(\tau)-v_{x}(\tau))^2 && \\ & + \frac{w_{d-lat}}{(\Delta d_{\sigma+\alpha}^{NV}(\tau))^2+\varepsilon^2}, \nonumber
        \end{align}
        \begin{align}\label{}
            &\Delta d_{\sigma+\alpha}^{NV}(\tau) \\ \nonumber
            & = \sqrt{(p_{x,\sigma+\alpha}^{NV}(\tau)-p_{x}(\tau))^2+(p_{y,\sigma+\alpha}^{NV}(\tau)-p_{y}(\tau))^2} - l,
        \end{align}
        \begin{align}\label{eq:lat-sgn}
            \xi_{\sigma+\alpha}^{NV}(\tau)  = \left\{ \begin{aligned}
                &1\quad\quad \text{if} \ v_{x,\sigma+\alpha}^{NV}(\tau) - v_{x}(\tau) >0\\
                &0 \quad\quad \text{if} \ v_{x,\sigma+\alpha}^{NV}(\tau) - v_{x}(\tau) \leq0,\\
            \end{aligned}
            \right.
        \end{align}		
        where $ v_{x,\sigma+\alpha}^{NV}$ denotes the velocity of the NV on the target lane; 
        $p_{x,\sigma+\alpha}^{NV}$ and $p_{y,\sigma+\alpha}^{NV}$ are the positions of the NV on the target lane; 
        the positive constants $w_{v-lat}$ and $w_{d-lat}$ are weighting coefficients; 
        $\xi^{NV}$ is a switching function to only account for the penalty on velocity differences if the NV is faster than the EV.
        By introducing penalties arising from rapidly approaching vehicles from behind on the adjacent lane, the safety in a lane-changing process is further enhanced. Therefore, we have the soft cost considering the NVs as follows: 
        \begin{align} \label{eq:l_bxu}
           \ell_{\tau,s\_NV} = b_{-1}(\tau) \ell_{d-lat, -1}(\tau) + b_{1}(\tau) \ell_{d-lat, 1}(\tau). \yesnumber 
        \end{align}

        With the inclusion of the collision avoidance constraint as the soft cost, the objective function in the first stage is defined as follows: 
        \begin{align} \label{}
           & \ell_\tau\big(b(\tau),\bar{x}(\tau),\bar{u}(\tau)\big) \\
           & = \ell_{\tau, tr} + \ell_{\tau, eff} + \ell_{\tau, rc} + \ell_{\tau, s\_LV} + \ell_{\tau,s\_NV}.   \nonumber
        \end{align}

        \subsubsection{Big-M Method}
            The big-M formulation is utilized to transform the logical implication involved in \eqref{eq:lon-sgn} and \eqref{eq:lat-sgn}. 
            With a sufficiently large $M \in \mathbb{R}^+$, the corresponding mixed integer constraints for \eqref{eq:lon-sgn} are derived as follows:
             \begin{align} 
                 & v_{\sigma+\alpha}^{LV}(\tau) - v_x(\tau) \geq -M \xi_{\sigma+\alpha}^{LV}(\tau), \label{eq:M-LV1} \\
                 & v_{\sigma+\alpha}^{LV}(\tau) - v_x(\tau) \leq M(1- \xi_{\sigma+\alpha}^{LV}(\tau)) - \epsilon, \label{eq:M-LV2}
             \end{align}
             and similarly, the mixed integer constraints regarding \eqref{eq:lat-sgn} are derived as:
             \begin{align} 
                 & v_{\sigma+\alpha}^{NV}(\tau) - v_x(\tau) \geq \epsilon - M (1-\xi_{\sigma+\alpha}^{NV}(\tau)), \label{eq:M-NV1}\\
                 & v_{\sigma+\alpha}^{NV}(\tau) - v_x(\tau) \leq M\xi_{\sigma+\alpha}^{NV}(\tau). \label{eq:M-NV2}
             \end{align}
			
        \subsubsection{Mixed Integer Programming} \label{sec:mip_formulation}
            With the above configurations, the MIP problem for a given prediction horizon $T$ is formulated as follows: 
            \begin{subequations} 
            \begin{align} 
    		      \displaystyle\operatorname*{minimize}_{b(\tau),\bar{x}(\tau),\bar{u}(\tau)}\quad
    			&   \sum_{\tau=0}^{T-1}\ell_\tau\big(b(\tau),\bar{x}(\tau),\bar{u}(\tau)\big),   \label{eq:mip_obj} \\
    			\operatorname*{subject\ to}\quad
                & \bar{x}(\tau+1) = F(\bar{x}(\tau), \bar{u}(\tau)), \label{eq:mip_vm} \\
                & v_x \ge \rho \mid{v_y}\mid, v_{x,\text{min}} \geq 0, \\
                & v_{\sigma+\alpha}^{LV}(\tau) - v_x(\tau) \geq -M \xi_{\sigma+\alpha}^{LV}(\tau),  \label{eq:bm0} \\
                & v_{\sigma+\alpha}^{LV}(\tau) - v_x(\tau)  \\ & \quad \leq M(1- \xi_{\sigma+\alpha}^{LV}(\tau)) - \epsilon, \nonumber \\
                &  v_{\sigma+\alpha}^{NV}(\tau) - v_x(\tau) \\ & \quad \geq \epsilon - M (1-\xi_{\sigma+\alpha}^{NV}(\tau)), \nonumber \\
                & v_{\sigma+\alpha}^{NV}(\tau) - v_x(\tau) \leq M\xi_{\sigma+\alpha}^{NV}(\tau), \label{eq:bm_end} \\
                & a_{x,\text{min}} \leq a_x(\tau) \leq a_{x,\text{max}},  \\
                & a_{y,\text{min}} \leq a_y(\tau) \leq a_{y,\text{max}}, \\
                & x(0)=x_0, \\
                & \sum_{\alpha}b_{\alpha}(\tau) = 1, \\
                &b_{\alpha}(\tau) \in \big\{0,1\}, \\ 
                &\alpha \in \{{-1},0,1\},   \label{eq:mip_sumB} \\
    			&\tau=0,1,\dotsm, T-1.  \nonumber
            \end{align}   
            \end{subequations}

            \begin{remark}
                The original problem formulation is nonlinear and nonconvex as noted in Section~\ref{sec:NLPP}. 
                The nonlinear bicycle model introduced in Section~\ref{sec:nonlinear_vm} is replaced by a linear vehicle model ~\eqref{eq:mip_vm}. The nonconvex collision avoidance constraint is also converted into soft penalty terms in the objective function ~\eqref{eq:mip_obj}.
                Basically, the nonlinear and nonconvex aspects in the original problem formulation are transformed through approximation and relaxation.
                Afterward, the optimal sequence of decisions can be obtained by solving the above MIP problem using the branch and bound algorithm{~\cite{land2010automatic}}, which approaches the optimal solution by dividing the solution space into branches and bounding the potential solutions.
                The solution from the MIP using ~\eqref{eq:mip_obj} to ~\eqref{eq:mip_sumB} can provide informed initialization for the original nonlinear and nonconvex problem.
            \end{remark}

        \subsection{Trajectory Planning Optimization} \label{sec:ilqr}
            With the outputs from the first stage, the optimal sequence of decisions is determined, which directly dictates the coordinates of the reference lane, denoted as $p^r_{x}$ and $p^r_{y}$, and the reference velocity, denoted as $v^r_{x}$.
            With these inputs, the optimization problem in the second stage narrows down to a general trajectory planning problem.
            The objective of the second stage is to generate a feasible trajectory based on the optimal sequence of decisions from the first stage.
        
        \subsubsection{Cost Function}
            The objective function in the second stage is the same as \eqref{eq:2a} in the nonlinear programming problem. 
            Since the binary variable $b$ is determined by solving the MIP problem in the first stage,  the objective function in each timestamp is constructed as follows:
            \begin{IEEEeqnarray}{rCl}~\label{eq:mp_obj_exp}
            &&\ell_\tau\big(x(\tau),u(\tau)\big) \\ &=&
            q_1\Big\|p_{x}(\tau) - p^r_{x}(\tau) \Big\|^2 + q_2\Big\|p_{y}(\tau) - p^r_{y}(\tau) \Big\|^2 \nonumber \\
            && + q_3\Big\|v_{x}(\tau) - v^r_{x}(\tau) \Big\|^2 + r_1\Big\|\delta(\tau) \Big\|^2 + r_2\Big\|a(\tau) \Big\|^2, \nonumber
            \IEEEeqnarraynumspace
        	\end{IEEEeqnarray}   
            where $q_1$, $q_2$, $q_3$, $r_1$, and $r_2$ represent the weighting parameters of the X-coordinate position tracking error, Y-coordinate position tracking error, longitudinal velocity tracking error, steering wheel input, and acceleration, respectively.
        
        \subsubsection{Constraints}
            The control inputs are bounded due to the physical limitations of the vehicle.
            Formally, the constraints are as follows: 
            \begin{align}
                a_{\text{min}} &\leq a(\tau) \leq a_{\text{max}}, \label{eq:mp_a} \\
                \delta_{\text{min}} &\leq \delta(\tau) \leq \delta_{\text{max}}. \label{eq:mp_steer}
            \end{align}
            
            Note that $a_{\text{min}}$ and $a_{\text{max}}$ denote the minimum and maximum acceleration values, which are constrained by the braking force limit of the vehicle and the power limit of the engine. 
            The range between $\delta_{\text{min}}$ and $\delta_{\text{max}}$ represents the possible steering angles, constrained by the physical limits of the vehicle. 

            Next, the collision avoidance constraint \eqref{eq:2d} is considered. 
    	In this study, the EV is represented by a rectangle, and the SVs are represented as ellipses.
            Our formulation takes into account the dimensions of the SVs and their relative motion with the EV.
            Primarily, the length of the SVs is taken into consideration in the approximation of the ellipse, which is characterized by $l_a$ and $l_b$ (lengths of the major and minor axes).
            Additionally, if either the EV or the SVs are traveling at an elevated velocity, a larger safety distance should be considered for collision avoidance, which is represented by an increased value in $l_a$.
            This proactive approach augments the longitudinal distances to afford the EV additional space to modulate its velocity and execute collision avoidance. 

            The collision avoidance constraint is then formulated as:
            \begin{equation}
            \begin{split}
            &\frac{1}{(l_a^i)^2} \left[ (p_x(\tau) - p_x^{i}(\tau)) \cos A^{i}(\tau) \right.\\
            &\left. + (p_y(\tau) - p_y^{i}(\tau)) \sin A^{i}(\tau) \right]^2 \\
            &+ \frac{1}{(l_b^i)^2} \left[ (p_x(\tau) - p_x^{i}(\tau)) \sin A^{i}(\tau) \right.\\
            &\left. - (p_y(\tau) - p_y^{i}(\tau)) \cos A^{i}(\tau) \right]^2 \geq 1.
            \end{split}
            \end{equation}


            Note that $p_{x}(\tau)$ and $p_{y}(\tau)$ are the positions of the EV. The superscript $i$ is the index used to denote the SV. $p_{x}^{i}(\tau)$ and $p_{y}^{i}(\tau)$ are the positions of the SV. The constants $A^{i}$, $l_a^i$, and $l_b^i$ are the heading angle and lengths of the major and minor axes of the ellipse used to approximate the SV, respectively. 

        \subsubsection{Trajectory Planning with ADMM-Based CiLQR}
            The trajectory planning problem in the second stage is therefore defined as follows:
            \begin{IEEEeqnarray*}{cl}~\label{eq:mp}
                \displaystyle\operatorname*{minimize}_{x(\tau),u(\tau)}\quad
    			&   \sum_{\tau=0}^{T-1}\ell_\tau\big(x(\tau),u(\tau)\big), \yesnumber \\
                \operatorname*{subject\ to}\quad
                & x(\tau+1)=f\big(x(\tau),u(\tau)\big), \\
                & g(x(\tau))\ge0, \\
                & a_{\text{min}} \leq a(\tau) \leq a_{\text{max}},  \\
                & \delta_{\text{min}} \leq \delta(\tau) \leq \delta_{\text{max}}, \\
                & x(0)=x_0, \\
                &\tau=0,1,\dotsm, T-1.
            \end{IEEEeqnarray*}
            
            The above trajectory planning problem, incorporating nonlinear system dynamics, collision avoidance constraints, and control input constraints, is effectively addressed through the alternating direction method of multiplier (ADMM) based constrained iLQR algorithm~\cite{ma2022alternating}. This algorithm splits the motion planning problem into manageable sub-problems leveraging the ADMM, and
            iteratively applies the iLQR with projection to deal with all the constraints incurred. With this approach, the optimal control inputs and the corresponding trajectory are generated with high computational efficiency.

    \subsection{Discussion}
        Our effort in designing the decision making and trajectory planning modules as a local planner is represented by the nonlinear programming problem in Section~\ref{sec:NLPP}. 
        Essentially, the integrated objective function, which is constructed from the costs in common to the decision making and trajectory planning tasks, offers the basis for formulating the two modules into a constrained optimization problem. 
        The term ``local'' refers to the proposed planner's capability in fulfilling lane-level planning, including the reference lane selection and respective trajectory generation. 
        In dividing the optimization problem into two sub-stages, the first stage by the TSO-based approach works on the two optimizing variables concurrently so that both an informed decision and a heuristic trajectory are offered to the second stage as initialization. By formulating the first stage as a proxy for the ultimate solution and utilizing the initialization from the first stage, the two stages devised according to the TSO-based approach are optimized towards the improvement in the quality of the ultimate solution, thereby ensuring the ultimate solution from the local planner is coherent.

        Regarding the optimality of the solution, the global optimum can often be found and proven for the standard mixed integer linear programming (MILP) and convex mixed integer quadratic programming (MIQP), benefiting from a well-established theory. However, as the MIP involved in our study is nonlinear and nonconvex, claiming optimality is a challenging mission. Nevertheless, we utilize 
        a small enough optimality gap (i.e., the difference between the best found solution and the lower bound of the objective function) as standardized in the MIP solver to accept a solution in the first stage. Moreover, in the context of autonomous driving, we evaluate the quality of the solution from the perspectives of safety, efficiency, and comfort in the ultimate driving performance.

\begin{table*}[htbp]
  \centering
  \caption{The settings in the four representative scenarios}
  \resizebox{1.0\linewidth}{!}{
    \begin{tabular}{cccccccccccc}
    \toprule
    \multicolumn{1}{c}{\multirow{2}[3]{*}{Vehicle ID}} & \multicolumn{1}{c}{\multirow{2}[3]{*}{Vehicle Types}} & \multicolumn{2}{c}{Scenario 1} &       & \multicolumn{2}{c}{Scenario 2} &       & \multicolumn{2}{c}{Scenario 3} & \multicolumn{2}{c}{Scenario 4} \\
\cmidrule{3-4}\cmidrule{6-7}\cmidrule{9-12}          &       & Position  & Velocity &       & Position  & Velocity &       & Position  & Velocity & Position  & Velocity \\
    \midrule
    0     & EV    & (0 m, 4 m) & 8 m/s &       & (0 m, 4 m) & 8 m/s &       & (0 m, 4 m) & 8 m/s & (0 m, 4 m) & 8 m/s \\
    1     & LV on the left lane & (12 m, 8 m) & 10 m/s &       & (12 m, 8 m) & 10 m/s &       & (8 m, 8 m) & 6 m/s & (15 m, 8 m) & 10 m/s \\
    2     & LV on the same lane & (15 m, 4 m) & 12 m/s &       & (15 m, 4 m) & 4 m/s &       & (15 m, 4 m) & 4 m/s & (10 m, 4 m) & 6 m/s \\
    3     & LV on the right lane & (8 m, 0 m) & 6 m/s &       & (8 m, 0 m) & 6 m/s &       & (12 m, 0 m) & 15 m/s & (5 m, 0 m) & 6 m/s \\
    4     & NV on the left lane & -     & -     &       & -     & -     &       & (-3 m, 8 m) & 10 m/s & -     & - \\
    5     & NV on the right lane & -     & -     &       & -     & -     &       & (-5 m, 0 m) & 6 m/s & -     & - \\
    \bottomrule
    \end{tabular}%
    }\label{tab:specs}%
\end{table*}%
     
\section{Experimental Results} \label{sec:exp}
    In this section, we show that:
    \begin{enumerate}
        \item Our proposed local planner can generate a sequence of optimal decisions and the respective trajectory, with a direct improvement in driving performance in the aspects of safety and traveling efficiency (Section \ref{sec:scenarios}).
        \item The TSO-based approach is essential in providing a high-quality solution (i.e., enhancements in safety and traveling efficiency) when compared to the initialization with pre-defined reference lanes and the ablations of the TSO-based approach.
        (Section \ref{sec:ablation}).
        \item We provide a closed-loop simulation in CARLA to demonstrate the efficacy of the proposed planner to adapt to dynamic traffic with fast execution time (Section \ref{sec:carla}).
    \end{enumerate}

    \subsection{Simulation Settings}
        All the algorithms are implemented in Ubuntu 20.04 LTS using Intel(R) Xeon(R) W-2223 CPU @ 3.6GHZ, NVIDIA RTX A2000 GPU with 4\,GB RAM and 12\,GB graphical memory.
        For the TSO-based approach, the first stage involving an MIP problem is solved using Gurobi~\cite{gurobi}, while the second stage is formulated as a CiLQR problem and solved using the ADMM-based method provided by~\cite{ma2022alternating} using Python 3.10.
    
        We use $N = 10$ and $\Delta t = 0.5\,\text{s}$ to output an optimal sequence of lane selection decisions, 
        and $N = 50$ and $\Delta t = 0.1\,\text{s}$ to output a trajectory, so the prediction horizon of $5\,\text{s}$ is ensured for both the decision making and trajectory planning results.
        Note that this configuration aligns with the common practice~\cite{wang2020infusing} that the decision making module requires a lower updating frequency than the trajectory planning module.

    \subsection {Performance Analysis} \label{sec:scenarios}
        \subsubsection{Scenario Description}
            According to the hyper-parameters listed in Table ~\ref{tab:specs}, we present four scenarios based on a three-lane highway where multiple SVs are involved.
            In Scenario~1, the LV on the same lane is traveling faster than the EV. It is usually not advisable to change the lane in such a situation. 
            Scenario 1 is crafted to validate the EV's capability to produce an optimal decision outcome utilizing the proposed planner, which is intended to adhere to lane-keeping decisions and follows the preceding vehicle.      
            In Scenario~2, the EV encounters a slow LV. This situation is to examine the planner's ability to determine a correct decision regarding the lane-changing process. 
            Naturally, the EV can change to the adjacent lane and achieve a higher traveling velocity. Yet, it is still a composite decision whether to change to the left or right lane. 
            The influence of the NVs is considered in Scenario~3, and the EV needs to evaluate the safety distance with the NVs in the target lane in a lane change decision.
            Scenario~4 involves a more complicated decision making process. At first, the EV encounters a slow LV on the current traveling lane. After a decision of lane change, the EV again encounters a vehicle ahead and needs to reevaluate the decision to keep or change the lane. It is demonstrated that our proposed planner can output an optimal sequence of decisions, rather than adhering to one fixed decision for the entire prediction horizon.
        
        \subsubsection{Results and Discussions}
\begin{figure}[t]
        		\centering
        		\includegraphics[trim=0 0 0 0, clip, width=1\linewidth]{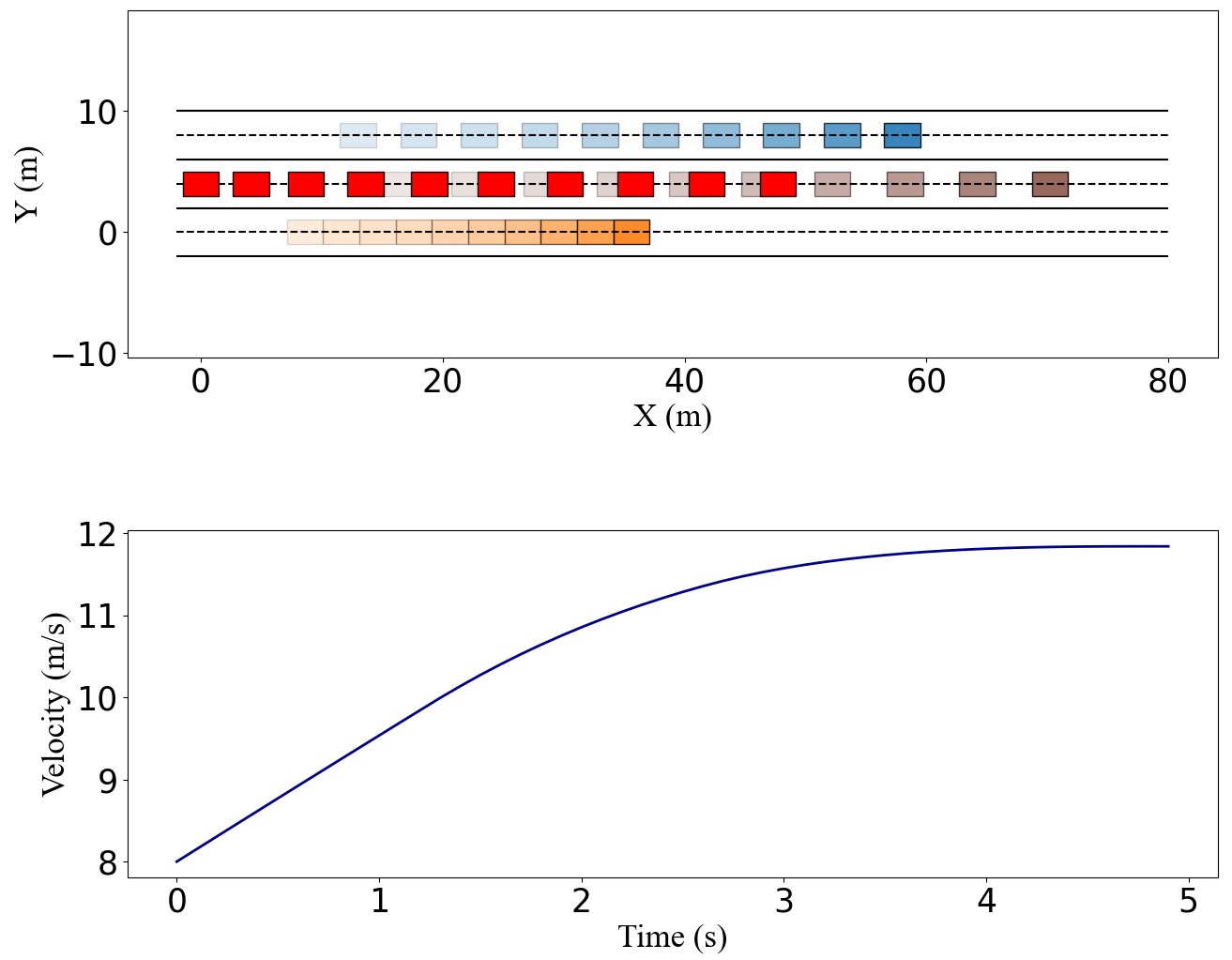}
        		\caption{Trajectory and longitudinal velocity of the EV in Scenario 1. Due to the fast-traveling LV on the same lane (in brown), the EV (in red) decides on lane keeping and smoothly brings up its velocity.}
        		\label{fig:case_1}
        	\end{figure}

            The experimental results in one prediction horizon of $5\,\text{s}$ for the four scenarios are demonstrated with visualizations of the trajectory and longitudinal velocity of the EV. 
            In Scenario 1 shown in Fig.~\ref{fig:case_1}, the EV tracks the current traveling lane because the decision result by the proposed planner is lane keeping. In the meanwhile, the velocity of the EV smoothly 
            increases from its initial velocity to $12\,\text{m/s}$, which is determined by the velocity of the fast LV.

\begin{figure}[t]
        		\centering
        		\includegraphics[trim=0 0 0 0, clip, width=1\linewidth]{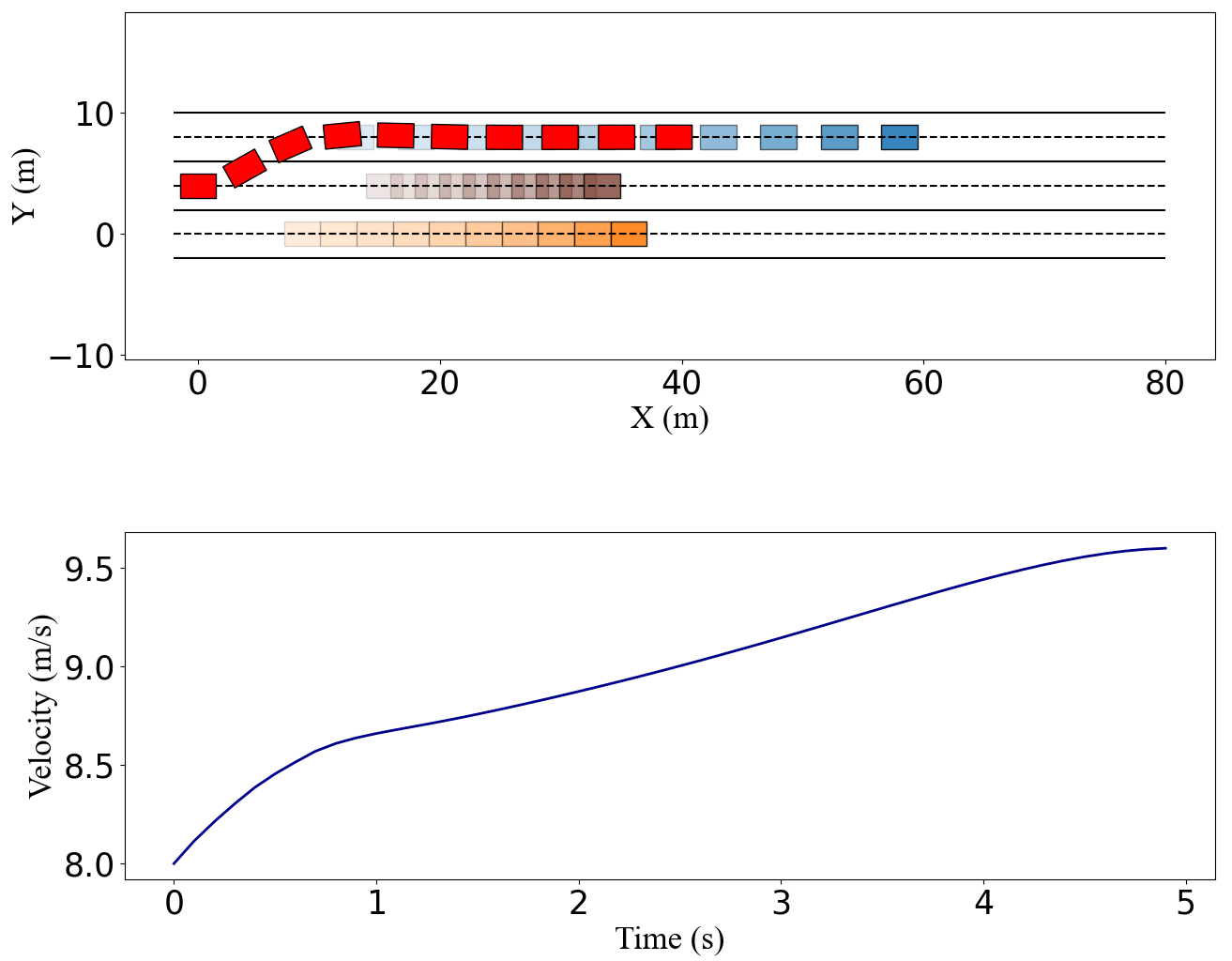}
        		\caption{Trajectory and longitudinal velocity of the EV in Scenario 2. The EV (in red) decides to change lanes to the left and generates a trajectory that completes this maneuver, because of the slow LV on the same lane (in brown) and the right lane (in orange). The velocity initially decreases due to the slow LV on the same lane and then increases to around $9.5\,\text{m/s}$ as determined by the velocity of the LV on the left lane (in blue).}
        		\label{fig:case_2}
        	\end{figure}

            In Scenario 2 shown in Fig.~\ref{fig:case_2}, the EV is faced with a slow-moving LV. 
            The decision result of our proposed planner is to change to the left lane. 
            Hence, the EV aims to track the adjacent left lane as the reference lane. The reference velocity is, in turn, given by the velocity of the LV on the left lane.
            The rationale behind the decision result is that the LV on the left lane travels at a higher velocity, and there is more longitudinal space available for the left lane change maneuver.

\begin{figure}[t]
        		\centering
        		\includegraphics[trim=0 0 0 0, clip, width=1\linewidth]{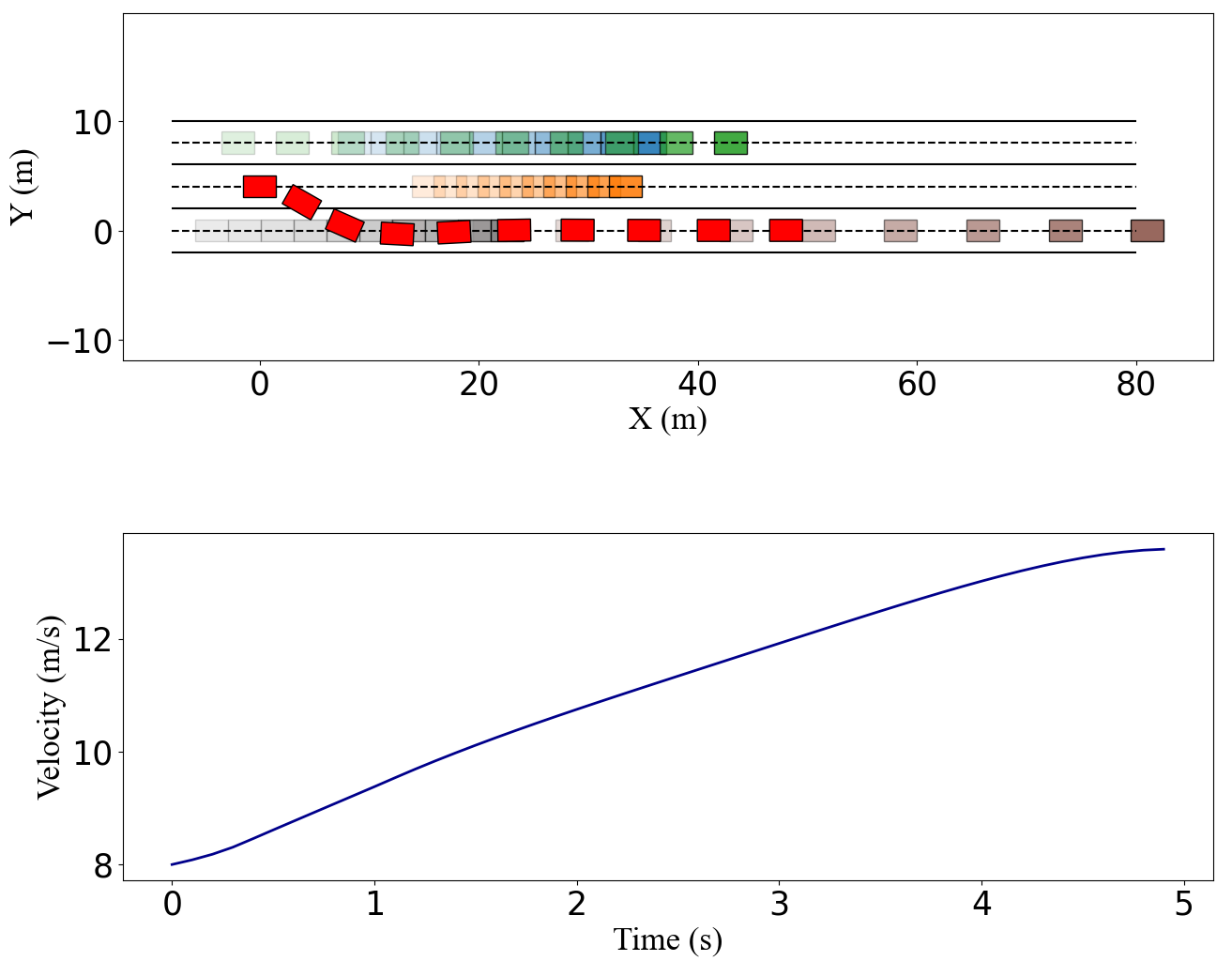}
        		\caption{Trajectory and longitudinal velocity of the EV in Scenario 3. The decision by the proposed planner is a right lane change, according to the slowly approaching NV (in grey) and the fast LV (in brown) on the right lane as compared to the traffic conditions on the left lane. The EV (in red) completes the right lane change maneuver with a smooth trajectory without collision and manages to increase its velocity to around $15\,\text{m/s}$ which is determined by the velocity of the LV on the right lane. }
        		\label{fig:case_3}
        	\end{figure}

            As illustrated in Fig.~\ref{fig:case_3}, in Scenario 3, the decision result given by our proposed planner is to change lanes to the right. 
            A major contribution to the decision result could be that the NV on the right lane is traveling relatively slowly, which provides more safety assurance.
            The EV can generate a smooth trajectory tracking the right lane and gradually increase its velocity according to the velocity of the LV on the right lane.

\begin{figure}[t]
        		\centering
        		\includegraphics[trim=0 0 0 0, clip, width=1\linewidth]{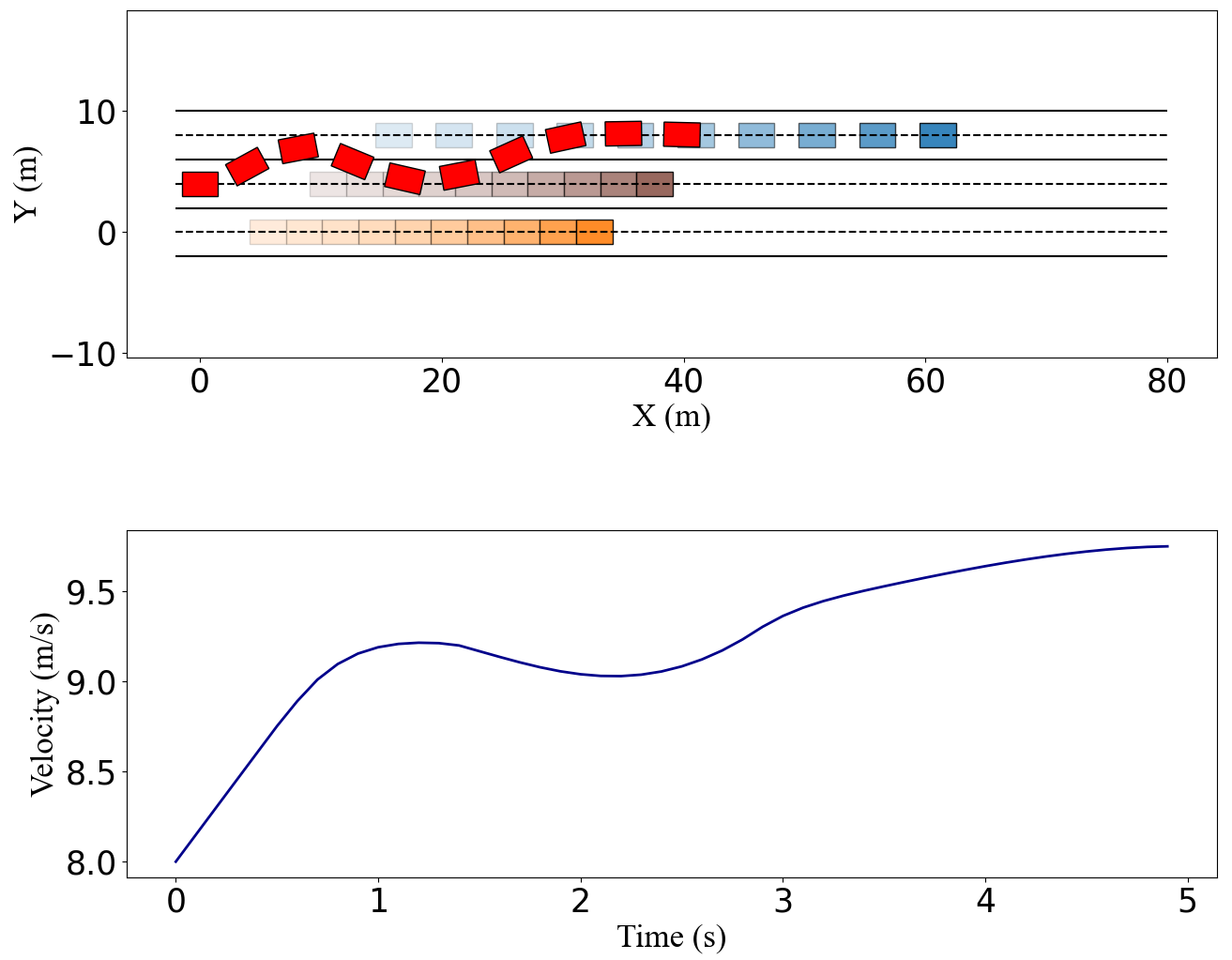}
        		\caption{Trajectory and longitudinal velocity of the EV in Scenario 4. The EV (in red) executes consecutive lane change maneuvers that lead to an overtaking and a left lane change with a safe and smooth trajectory. The velocity is first decreased as associated with the overtaking maneuver at around $2.5\,\text{s}$, but is then resumed to around $10\,\text{m/s}$ after the decision to change to the left lane at the end of the prediction horizon.}
        		\label{fig:case_4}
        	\end{figure}

           Consecutive lane changes are observed in Fig.~\ref{fig:case_4}, which in turn constitute an overtaking maneuver. In Scenario~4, the EV is at first faced with a slow LV on the current traveling lane and changes lanes to the left. After the completion of the lane change, the EV again encounters a vehicle ahead and reconsiders the decisions on lane keeping or lane changing.
           Due to the low velocity of the LV on the original lane, there is free space on the original lane so the EV chooses to change back to the original lane, which leads to an overtaking. 
           Following the overtaking, the decision making process is completed with a lane change to the left to pursue higher velocity.
           Through consecutive movements in the prediction horizon, the EV is able to adjust its longitudinal velocity to achieve higher traveling efficiency.
         
            In summary, our proposed planner is capable of optimizing for the decisions and a feasible trajectory that matches the scenario design in the above conditions.
        
    \subsection{Ablation Study of the TSO-Based Approach} \label{sec:ablation}
        We consider three categories of ablation studies. 
        \begin{enumerate}
            \item Ablation Study 1: To unveil the importance of the concurrent optimization of decision making and trajectory planning in the proposed local planner,
            we consider a set of trajectory planners named iLQR.KEEP, iLQR.LEFT and iLQR.RIGHT, in which the planners solely track a pre-defined current traveling lane, left adjacent lane, and right adjacent lane as the reference lane, respectively.
            \item Ablation Study 2: We investigate the importance of the high-fidelity vehicle model to the delivery of the ultimate driving result of the whole system. 
            A similar MIP formulation in the first stage is considered, without using the high-fidelity vehicle model in the second stage.
            \item Ablation Study 3: We evaluate the effects of transformation of collision avoidance into soft cost in the first stage optimization. An MIP problem without hard enforcement of collision avoidance is considered.
        \end{enumerate}
        
        The Ablation Study 1 is implemented using the ADMM-based CiLQR algorithm, and the Ablation Study 2 and 3 are put into MIP formulation and solved using Gurobi.

        The driving safety is measured in terms of the number of obstacles and collisions. 
        When the EV and the SVs are in close distance wherein the obstacle avoidance has to be evoked, the number of obstacles is increased by one. 
        As one major advantage of the proposed local planner is to allow the decision making stage to provide informed guidance to the trajectory planning problem.
        The less involvement in obstacle avoidance is an indication of improved safety.
        When an actual collision between the EV and the SVs happens, the number of collisions is increased by one.
        The traveling efficiency is represented by the longitudinal distance that the EV progresses during the prediction horizon, and its average velocity.   

        \begin{table}[htbp]
          \centering
          \caption{Comparison of driving performance in one prediction horizon}
          \resizebox{1.0\linewidth}{!}{
            \begin{tabular}{ccc|cccc}
            \toprule
                  &       &       & \multicolumn{2}{c}{(a) Driving Safety} & \multicolumn{2}{c}{(b) Traveling Efficiency} \\
                  &       &       & Obstacle & \multicolumn{1}{c}{Collision } & Progress (m) & Velocity (m/s) \\
            \midrule
            \multirow{6}[8]{*}{Scenario 1} & \multicolumn{1}{c}{\multirow{3}[2]{*}{Ablation Study 1}} & \multicolumn{1}{c|}{iLQR.KEEP} & 0     & 0     & 52.36 & 10.73±1.21 \\
                  &       & \multicolumn{1}{c|}{iLQR.LEFT} & 0     & 0     & 45.21 & 9.24±0.41 \\
                  &       & \multicolumn{1}{c|}{iLQR.RIGHT} & 0     & 0     & 45.34 & 9.33±0.47 \\
        \cmidrule{2-7}          & \multicolumn{1}{c}{Ablation Study 2} & \multicolumn{1}{c|}{-} & 0     & 0     & 35.98 & 6.32±0.77 \\
        \cmidrule{2-7}          & \multicolumn{1}{c}{Ablation Study 3} & \multicolumn{1}{c|}{-} & 0     & 0     & 52.36 & 10.73±1.21 \\
        \cmidrule{2-7}          & \multicolumn{2}{c|}{\textbf{OURS}} & \textbf{0} & \textbf{0} & \textbf{52.36} & \textbf{10.73±1.21} \\
            \midrule
            \multirow{6}[8]{*}{Scenario 2} & \multicolumn{1}{c}{\multirow{3}[2]{*}{Ablation Study 1}} & \multicolumn{1}{c|}{iLQR.KEEP} & 1     & 0     & 40.73 & 8.34±0.27 \\
                  &       & \multicolumn{1}{c|}{iLQR.LEFT} & 0     & 0     & 43.08 & 8.98±0.43 \\
                  &       & \multicolumn{1}{c|}{iLQR.RIGHT} & 1     & 0     & 38.9  & 8.10±0.08 \\
        \cmidrule{2-7}          & \multicolumn{1}{c}{Ablation Study 2} & \multicolumn{1}{c|}{-} & 0     & 2     & 35.97 & 8.19±1.19 \\
        \cmidrule{2-7}          & \multicolumn{1}{c}{Ablation Study 3} & \multicolumn{1}{c|}{-} & 1     & 0     & 40.73 & 8.34±0.27 \\
        \cmidrule{2-7}          & \multicolumn{2}{c|}{\textbf{OURS}} & \textbf{0} & \textbf{0} & \textbf{43.08} & \textbf{8.98±0.43} \\
            \midrule
            \multirow{6}[8]{*}{Scenario 3} & \multicolumn{1}{c}{\multirow{3}[2]{*}{Ablation Study 1}} & \multicolumn{1}{c|}{iLQR.KEEP} & 1     & 0     & 39.28 & 8.06±0.05 \\
                  &       & \multicolumn{1}{c|}{iLQR.LEFT} & 1     & 0     & 39.17 & 8.09±0.08 \\
                  &       & \multicolumn{1}{c|}{iLQR.RIGHT} & 0     & 0     & 53.25 & 11.11±1.71 \\
        \cmidrule{2-7}          & \multicolumn{1}{c}{Ablation Study 2} & \multicolumn{1}{c|}{-} & 0     & 1     & 36.32 & 5.14±1.86 \\
        \cmidrule{2-7}          & \multicolumn{1}{c}{Ablation Study 3} & \multicolumn{1}{c|}{-} & 1     & 0     & 39.28 & 8.06±0.05 \\
        \cmidrule{2-7}          & \multicolumn{2}{c|}{\textbf{OURS}} & \textbf{0} & \textbf{0} & \textbf{53.25} & \textbf{11.11±1.71} \\
            \midrule
            \multirow{6}[8]{*}{Scenario 4} & \multicolumn{1}{c}{\multirow{3}[2]{*}{Ablation Study 1}} & \multicolumn{1}{c|}{iLQR.KEEP} & 1     & 0     & 40.58 & 8.34±0.44 \\
                  &       & \multicolumn{1}{c|}{iLQR.LEFT} & 0     & 0     & 43.15 & 8.98±0.41 \\
                  &       & \multicolumn{1}{c|}{iLQR.RIGHT} & 1     & 0     & 39.44 & 8.19±0.07 \\
        \cmidrule{2-7}          & \multicolumn{1}{c}{Ablation Study 2} & \multicolumn{1}{c|}{-} & 0     & 1     & 36    & 4.00±0.00 \\
        \cmidrule{2-7}          & \multicolumn{1}{c}{Ablation Study 3} & \multicolumn{1}{c|}{-} & 1     & 0     & 40.58 & 8.34±0.44 \\
        \cmidrule{2-7}          & \multicolumn{2}{c|}{\textbf{OURS}} & \textbf{0} & \textbf{0} & \textbf{43.75} & \textbf{9.22±0.41} \\
            \bottomrule
            \end{tabular}%
          }\label{tab:ablation}%
        \end{table}%
 
        The metrics in driving safety and traveling efficiency with our proposed planner and the ablation methods in all four scenarios are summarized in Table~\ref{tab:ablation}.
        As shown in column (a), our proposed local planner solved by the TSO-based approach exhibits better driving safety. 
        It is observed that the methods in Ablation Study 1 are not able to navigate towards lanes with better driving conditions so obstacle avoidance has to be conducted.
        In Ablation Study 2, the safety constraints in Scenarios 2-4 are severely violated where actual collisions are recorded. 
        Without factoring in the interactions with the SVs in the first stage optimization (i.e., Ablation Study 3), the EV is not able to make appropriate decisions and sticks with lane keeping, and it experiences similar concerns as in Ablation Study 1 where obstacle avoidance is inevitable.

        In column (b), we observe that the traveling efficiency of our proposed method outperforms other methods.
        This underscores the benefit of the proposed local planner in improving driving conditions for the EV, consequently enhancing overall driving performance. The effects of trailing a slow-moving LV or encountering  SVs not only bring about safety concerns, but also diminish traveling efficiency. The collision avoidance maneuvers inherently mandate reduced velocity, thereby restricting the EV's longitudinal traversal. In the event of an actual collision, both travel distance and velocity are impacted.

        In summary, the proposed planner generates a sequence of optimal decisions and the corresponding trajectory for the EV, exhibiting direct improvement in driving performance.
        Driving safety is enhanced because the proposed planner can guide the EV towards the lane with lower chances of being blocked by the SVs. 
        The performance in traveling efficiency is also notably improved in the sense that the EV is able to traverse longer longitudinal distances with higher velocities.

    \subsection {Comparative Study by Closed-Loop Simulation} \label{sec:carla}
        \subsubsection {Simulation Settings}
        We compare the proposed planner with the following baselines to evaluate the efficacy of the proposed method: 1) FSM+PID: a pipeline where the rule-based decision making and PID-based trajectory planning are separately designed. 2) IDP: an integrated decision making and motion planning approach as proposed in~\cite{hang2020integrated}.
        The proposed planner and the baseline methods are deployed in a closed-loop simulation using CARLA~\cite{dosovitskiy2017carla}.
        CARLA provides a simulated driving environment on map information (i.e., reference lanes) and SVs (i.e., both static and in motion), operating at a fixed simulation frequency of $20$ Hz.
        In this closed-loop simulation featuring a sequentially connected scenario, the mentioned methods are implemented in a receding horizon fashion, with a planning step of $N_{p}=10$ and an executing step of $N_{c}=1$.
        To reflect driving performance in the entire simulation, where multiple predictions are made compared to the results for a prediction horizon in Section~\ref{sec:scenarios}, quantitative indices of safety, efficiency, and comfort~\cite{wu2023integrated} are introduced.

        The safety index is defined as follows:
        \begin{IEEEeqnarray}{rCl} \label{eq:safe_index}
            I_\text{safe} = \min_{i} S_i, \IEEEeqnarraynumspace
        \end{IEEEeqnarray}
        where
        \begin{IEEEeqnarray}{rCl} \label{eq:safe_index}
            S_i = \frac{\min(\Delta x_i,60)}{v_x}. \IEEEeqnarraynumspace
        \end{IEEEeqnarray}

        For each SV with index $i$, we calculate a safety score $S_i$ from the relative distance and current velocity of the EV. The safety score is $0$ if a collision occurs. The safety index is the minimum value considering the safety scores for all SVs. The safety index directly reflects the safety as higher value means larger safety spacing between the EV and SVs based on the EV's current velocity, and the index is zero if the EV crashes with SVs.

        The efficiency index, which compares the current velocity with the highest attainable velocity, is defined as follows:
        \begin{IEEEeqnarray}{rCl} \label{eq:eff_index}
            I_\text{eff} = 10\tanh \frac{1.83v_x}{\min(v^\text{limit}, v^{LV})}, \IEEEeqnarraynumspace
        \end{IEEEeqnarray}
        where $v^\text{limit}$ is the velocity limit; $v^\text{LV}$ and $v_x$ refer to the velocities of the LV and EV.
        With the use of the $\tanh$ function, the value of $I_\text{eff}$ is within the range of $0$ to $10$. 
        A higher efficiency index indicates a higher travel efficiency.

        The comfort index is measured by the weighted root-mean-square acceleration (WRMSA) as the control in acceleration is directly sensed by passengers onboard, which is defined as follows:
        \begin{equation}
            \text{{WRMSA}} = \sqrt{\frac{{\sum_{i=1}^{N} w_i \cdot a_i^2}}{{\sum_{i=1}^{N} w_i}}},
        \end{equation}
        where $N$ represents the total number of data points, \(w_i\) represents the weight associated with each data point, and $a_i$ represents the acceleration value for each data point.
        A lower value of the WRMSA is a reflection of a more comfortable riding experience.

        \subsubsection{Results and Discussions} \label{sec:carla_result}   
            \begin{figure*}[t]
        		\centering
        		\includegraphics[trim=0 0 0 0, clip, width=1\linewidth]{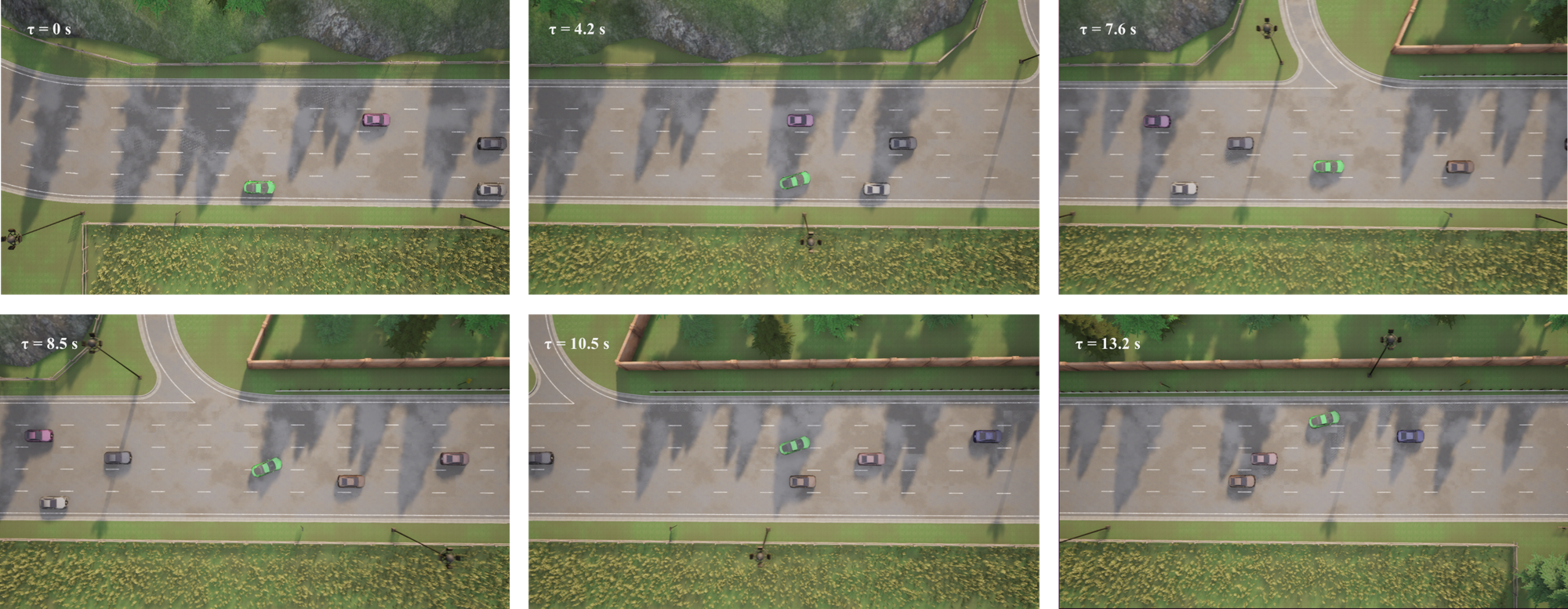}
        		\caption{Simulation results on the multi-lane urban driving scenarios in CARLA. Six keyframes are selected to demonstrate the optimal lane-selection decisions for the EV (in green) in the presence of SVs. Video demonstrations is available at ~\href{https://youtu.be/aDyPBBs3sOg}{https://youtu.be/aDyPBBs3sOg}.}
        		\label{fig:carla_scene}
        	\end{figure*}
            
            The five-lane highway scenario in CARLA is considered, with the EV initially positioned at the rightmost lane and six SVs present. Using the proposed planner, the EV is able to navigate through the dynamic scene with multiple nearby vehicles and it reaches the leftmost lane which is unoccupied by the SVs.

            Besides the lane-keeping decision, notably, there are four lane-change decisions in total along the planned trajectory. The selected simulation frames from CARLA are provided in Fig.~\ref{fig:carla_scene}.    
            The first lane-change decision takes place in round time of $4\,\text{s}$. The EV initially occupies the rightmost lane with its velocity accelerating from zero to its desired level of 13\,m/s. Then the EV faces the slower LV and gradually decreases its velocity as capped by the velocity of the LV. The EV equipped with our proposed planner starts a lane change to the adjacent left lane, which is the closest available lane considering the right boundary of the current lane. This lane-change decision enables the EV to raise its velocity up to 16.6\,m/s which is designated by the velocity limit.
            At around the time of $7\,\text{s}$, the EV encounters an LV whose velocity is at 6\,m/s. The decrease in the EV's velocity is observed due to the encountering of the slow-moving LV. In the meanwhile, there are NVs on the left lane. Considering that the NV is traveling at a lower velocity than the EV, the EV executes the lane-change decision by the planner and brings the velocity back to about 13\,m/s. 
            Subsequently, the EV finds itself trailing again behind a slow-moving LV and the decision process is more complex. The EV is given the options including lane keeping, left lane change, and right lane change.
            In addition to the LV on the same lane, there is an LV on the right lane and an NV on the left lane. 
            The EV is provided with a decision to change to the left lane by the proposed planner at around the time of $10\,\text{s}$.
            Finally, at around the time of $12\,\text{s}$, the EV chooses to go to the left lane in a situation where the current lane and the adjacent right lane are both occupied. 
            At the end of the simulation, the EV manages to navigate through this dynamic scenario occupied with multiple SVs from the right-most lane to the left-most lane where no SV is present. 
            During the whole simulation, the EV adjusts its velocity smoothly. The EV is able to maintain a high traveling efficiency, and resume its velocity after the slow-downs due to the encountering of the slow LVs and the lane-change maneuvers.
      
\begin{figure}[t]
            \centering
            \includegraphics[trim=0 0 0 0, clip, width=1\linewidth]{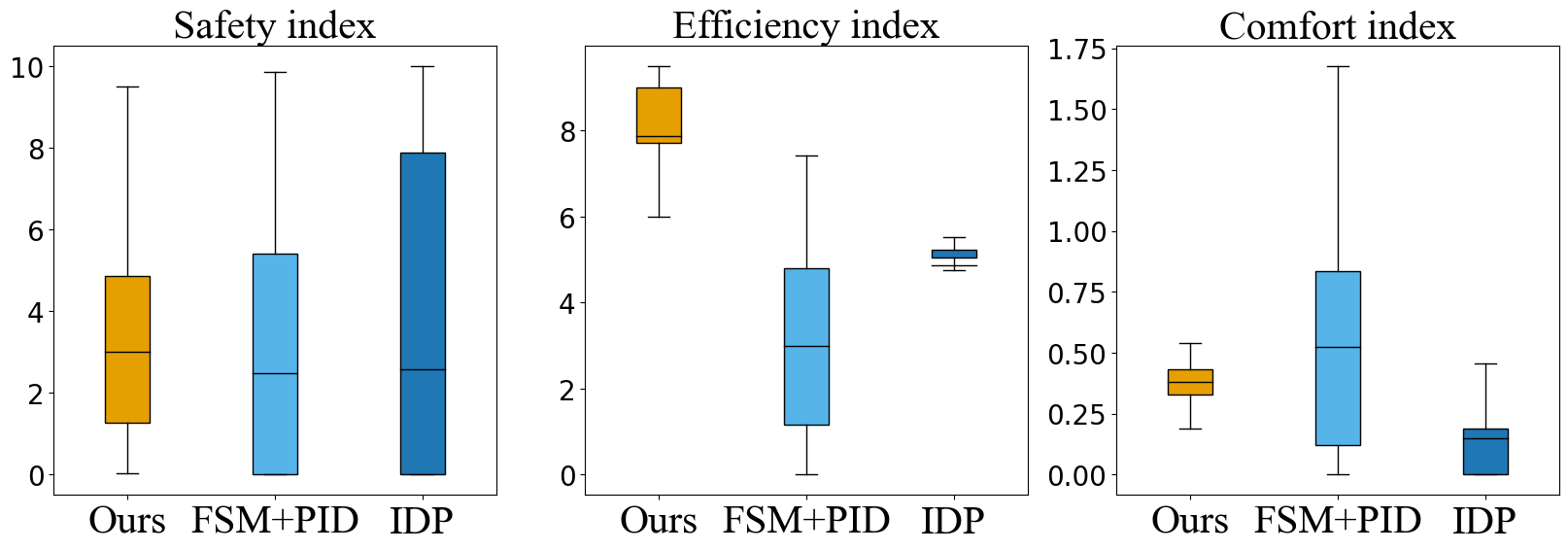}
            \caption{Comparison of driving performance in terms of safety, efficiency, and comfort indices between our proposed planner and two baseline methods in CARLA.}
            \label{fig:carla_index}
            \end{figure}

            The comparative results on safety index, efficiency index, and comfort index, with box plots indicating the respective performance are shown in Fig.~\ref{fig:carla_index}.
            The proposed planner is capable of passing the entire simulation without collision or hazardous behavior which is counted when the EV is traveling in close distance with the SVs and is of high risk of collisions. In contrast, \texttt{FSM+PID} method produces one collision and two occurrences of hazardous behaviors. The difference in terms of safety is reflected by the safe index: 
            The average safe index reported by the proposed planner, \texttt{FSM+PID} and \texttt{IDP} methods are $2.99$, $2.47$ and $2.58$, respectively. 
            A significant enhancement in the traveling efficiency is also observed for the proposed planner.
            The average efficiency index of the proposed planner is $7.88$, as compared with an average of $2.99$ for the \texttt{FSM+PID} method, and $4.87$ for the \texttt{IDP} method.
            Using the results from the proposed planner, the EV manages to navigate through the driving scenario with multiple dynamic vehicles with an average velocity of 11\,m/s, which underscores the planner's ability in attaining higher velocity with active lane selection decisions.
            The comfort index of the proposed planner is also lower than the \texttt{FSM+PID} method. Although the decision of lane changes by the proposed planner leads to variations in the lateral accelerations, the lane keeping decisions and the respective control inputs by \texttt{FSM+PID} method exhibit larger fluctuations in the longitudinal accelerations, which affects the comfort level adversely.

            Given that the evaluation thus far focused on low to medium velocity scenarios, we extend our analysis by conducting additional testing under high-velocity conditions in the CARLA simulator, with the aim to validate the robustness of the proposed planner across a broader range of operational speeds. The velocity limit is set at 90\,km/h, or 25\,m/s. The same initial positions of the SVs are followed, and the velocity for each SV is doubled to suit the high-velocity scenario. Our proposed planner is able to pass the testing with no collision or hazardous driving behaviors, and manages to increase its velocity from zero to as high as 23.42\,m/s, approaching the designated velocity limit of 25\,m/s in this testing scenario. The EV's velocity during the whole simulation is reported as 16.38\,m/s on average.

            Overall, the simulation results demonstrate the effectiveness of our proposed planner in tackling challenging urban autonomous driving tasks. 
            The improvements in our method mainly come from two perspectives: 
            1) the ability to simultaneously optimize for decisions and trajectories provides the EV with more flexibility in navigating towards lanes with more favorable conditions (i.e., less obstacles);
            and 2) the integrated objective on safety, efficiency, and comfort, aligns the explicit decision making and trajectory planning tasks with broader objectives in driving, thereby elevating the overall quality of performance.

        \subsubsection {Computational Efficiency Analysis} \label{sec:time}
            We provide an analysis on the performance of computational efficiency using the \texttt{IDP} method as a baseline. The execution time, which refers to the time consumed to execute one step in Carla, is shown in Table~\ref{tab:mpc_time_compare}.
            The average execution time is $0.12\,\text{s}$ for the proposed planner, and $0.15\,\text{s}$ for the baseline \texttt{IDP} method.
            Notably, the proposed planner can maintain stable computational efficiency with increased complexity in situations where there are SVs nearby and the interactions with the SVs are considered as part of the optimization process.
            \begin{table}[t]
              \centering
              \caption{Execution time of the proposed planner and comparison with baseline method in CARLA simulation}
                \begin{tabular}{c|cccc}
                \toprule
                \multirow{2}[4]{*}{Method} & \multicolumn{4}{c}{Time step} \\
            \cmidrule{2-5}          & 0-4 s & 4-8 s & 8-12 s & 12-16 s \\
                \midrule
                Ours  & 0.12±0.02 s & 0.12±0.02 s & 0.13±0.02 s & 0.13±0.02 s \\
                IDP   & 0.05±0.01 s & 0.13±0.15 s & 0.19±0.19 s & 0.17±0.18 s \\
                \bottomrule
                \end{tabular}%
              \label{tab:mpc_time_compare}%
            \end{table}%


\section{Conclusion} \label{sec:con}
   This paper presents a coherent local planner to generate the optimal sequence of decisions and the respective trajectory. The decision making and trajectory planning modules are jointly designed leveraging a constrained optimization formulation, with an integrated objective function that seamlessly integrates the discrete decision variables into the continuous trajectory optimization. The formulated MIP problem with nonlinearity and nonconvexity is solved using a TSO-based approach, with the key notion of formulating the two optimization stages using similar problem representations, to ensure the coherence of the results of decision making and trajectory planning in the local planner. We conduct exhaustive simulation trials in different traffic scenarios, and the results substantiate the efficacy of the proposed planner in accomplishing the decision making and trajectory planning tasks. The ablation studies show the effectiveness of the MIP-based first stage for the local planner, while the simulation in CARLA demonstrates a safe, efficient, and comfortable driving performance using the proposed planner.
   One limitation of this study is that the prediction task is not addressed. In the current settings, the predictions on SVs are provided as part of the input data. Hence, part of our future research endeavors is to incorporate the prediction tasks within the framework of this synergistic approach to make more informed decisions exhibiting enhanced interactivity between the EV and others.

\bibliographystyle{IEEEtran}
\bibliography{mybibfile}

\end{document}